\documentclass{article}
\usepackage{float}

\PassOptionsToPackage{numbers}{natbib}

\usepackage[eandd,final]{neurips_2026}

\usepackage[utf8]{inputenc} 
\usepackage[T1]{fontenc}    
\usepackage{textcomp} 
\usepackage{hyperref}       
\usepackage{url}            
\usepackage{booktabs}       
\usepackage{amsfonts}       
\usepackage{nicefrac}       
\usepackage{microtype}      
\usepackage[dvipsnames,table]{xcolor}
\usepackage{pifont}
\usepackage{xpatch}
\usepackage{tabularx}
\usepackage{fontawesome5}
\usepackage{colortbl}
\usepackage{amssymb}
\usepackage{tikz}
\usepackage{multirow}
\usepackage{xpatch}
\usepackage{caption}
\usepackage{tcolorbox}
\usepackage{tikz}
\usetikzlibrary{positioning, fit, arrows.meta, shapes.geometric}
\usepackage{amsmath, amssymb}
\usepackage{graphicx}
\usepackage{pifont}
\makeatletter
\xapptocmd{\NAT@bibsetnum}{\setlength{\leftmargin}{0pt}\setlength{\itemindent}{\labelwidth}\addtolength{\itemindent}{\labelsep}}{}{}
\makeatother

\newcommand{\cmark}{\textcolor{green!60!black}{\ding{51}}} 
\newcommand{\xmark}{\textcolor{red!80!black}{\ding{55}}}

\title{\textsc{PhysElite}: 
How Far Are LLMs from Solving Olympiad-Level Physics Problems?}

\author{%
  Ruoran Xu\textsuperscript{* \textdaggerdbl} \quad
  Wending Gao\textsuperscript{*} \quad
  Liyunfeng Chen\textsuperscript{*} \quad
  Aixin Shi\textsuperscript{*} \\
  Haoyu Cheng \quad
  Zixiang Fang \quad
  Yiqiang Zou \quad
  Qiufeng Wang\textsuperscript{\textdagger}\\
  Xi'an Jiaotong-Liverpool University
}

\begin{document}

\maketitle

\begingroup
\renewcommand{\thefootnote}{}
\footnotetext{
\textsuperscript{*}Equal contribution \quad
\textsuperscript{\textdaggerdbl}Project Lead \quad
\textsuperscript{\textdagger}Corresponding author\quad {qiufeng.wang@xjtlu.edu.cn}.}
\endgroup

\begin{abstract}
Understanding how (multimodal) large language models perform on physics problems requires benchmarks that reflect the difficulty and breadth of expert-level physical reasoning. Existing physics benchmarks remain limited in the following two important ways: (1) short of high-difficulty datasets, and (2) lack of comprehensive coverage of visual forms, knowledge points, and step-by-step solution processes. As a result, model performance on current datasets may not be fully representative of their ability to solve complex physics problems. To address these issues, we present \textsc{PhysElite}, a large-scale bilingual multimodal benchmark for Olympiad-level physics reasoning. \textsc{PhysElite} contains 11,586 Olympiad-tier problems. For each problem, we provide corresponding visual diagrams, step-by-step bilingual Chinese--English solution derivations, and the final answer. We benchmark 18 open-source and closed-source MLLMs, and find that even the strongest model reaches only 33.7\% answer accuracy. We additionally conduct step-level process evaluation to diagnose where models fail in the reasoning chain. Our code and dataset are released at \url{https://github.com/PremiLab-Math/PhysElite}.

\end{abstract}

\begin{figure}[h]
    \centering
    \includegraphics[width=0.9\linewidth]{firstfig.pdf}
    \caption{Comparision of two existing K-12 level samples and one illustrative example of olympiad-level physics problem with detailed multimodal annotation of solution steps in our \textsc{PhysElite} (The cooresponding completed example can be found in Figure~\ref{fig:completesample})} 
    \label{fig:placeholder1}
\end{figure}

\clearpage

\section{Introduction}

\begin{figure}[h]
    \centering
    \includegraphics[width=0.9\linewidth]{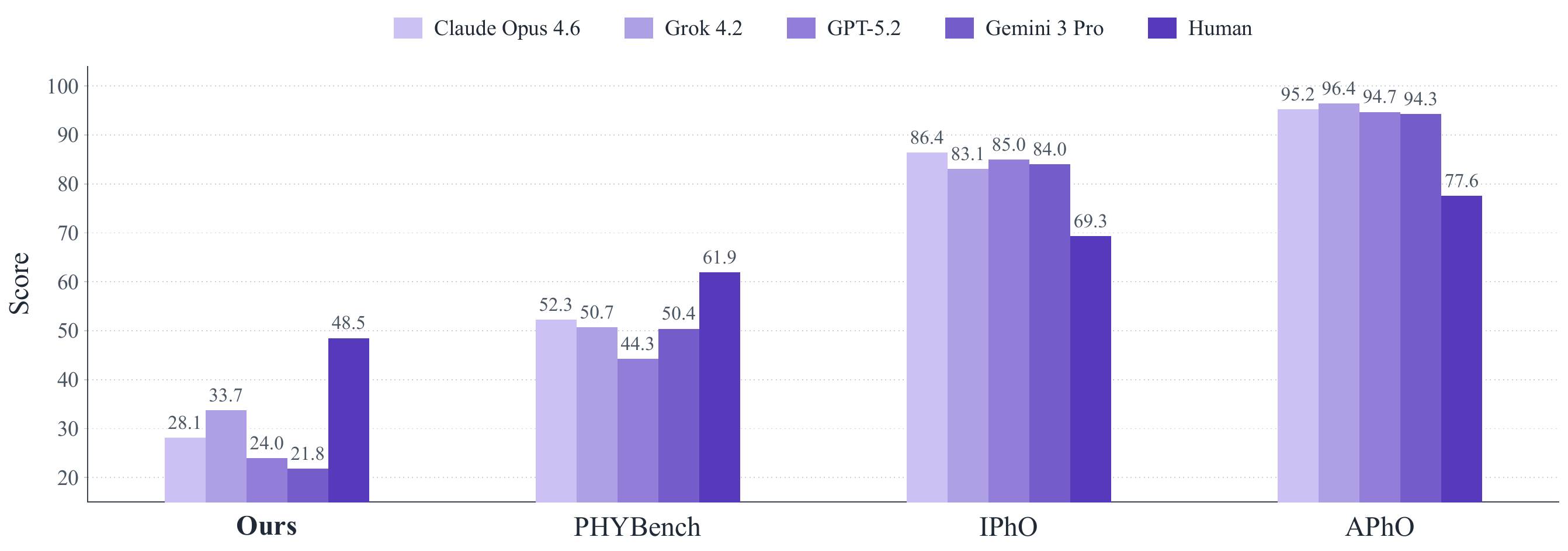}
    \caption{Performance of four state-of-the-art MLLMs on our \textsc{PhysElite} and three representative existing physics benchmarks: while prior benchmarks are nearing saturation, \textsc{PhysElite} leaves a substantial gap between the best model and the human reference.}
    \label{fig:performance}
\end{figure}

Physics problem-solving has long been a rigorous benchmark for (multimodal) large language models (MLLMs/LLMs), requiring models to parse structured visual input, apply physical laws under explicit constraints, and chain intermediate results into coherent multi-step derivations. In recent years, MLLMs/LLMs have achieved impressive performance on physics reasoning benchmarks~\citep{yu2025hipho,physunibench2025,xu2025ugphysics,qiu2025phybench}, with some even claiming medal-level results on Olympiad physics tasks~\citep{he-etal-2024-olympiadbench,yu2025hipho,chen2025p1}. However, these high scores raise a critical question: Are current models truly capable of solving authentic Olympiad-level physics problems?
To answer this question, we systematically analyze existing physics benchmarks and identify two core limitations:

\textbf{Scale and difficulty.} Most benchmarks are lack of challenging problems. Many existing datasets are dominated by elementary or K-12 level problems, which fail to stress-test frontier models’ ability to handle complex, multi-step reasoning. As shown in the examples in Figure~\ref{fig:placeholder1}, these problems lack the depth and multimodal complexity of real Olympiad tasks. On the otherhand, Datasets such as IPhO~\citep{ipho_unofficial} and APhO ~\citep{apho} represent Olympiad-level, but both contain only a limited number of problems and are updated infrequently, which make them vulnerable to model saturation, where performance plateaus before reaching the true ceiling of Olympiad problem-solving ability, as illustrated in Figure~\ref{fig:performance}

\textbf{Missing process-level annotations.} Many benchmarks provide only final answers, with no intermediate derivations linking visual cues to physical laws. For physics, the derivation itself carries more evaluative signal than the final answer, and grading only the final answer leaves the source of a model's error unclear. The absence of step-level annotations also limits the use of physics problems as a training signal for chain-of-thought supervision or process-level reinforcement learning, a direction that has already shown strong returns in many domains~\citep{lightman2024lets}.

To address these gaps, we introduce PHYSELITE, a large-scale bilingual multimodal benchmark for Olympiad-level physics reasoning. As summarized in Table 1, PHYSELITE contains 11,586 Olympiad-tier problems, each paired with visual diagrams, step-by-step bilingual derivations, and verified final answers (Figure~\ref{fig:placeholder1}). Our evaluation of 18 open-source and closed-source MLLMs on PHYSELITE reveals that even the strongest model achieves only 33.7\% answer accuracy, leaving a substantial gap with human experts (Figure~\ref{fig:performance}). We further conduct process-level evaluation to diagnose where models fail in the reasoning chain, identifying key failure modes that point to clear directions for future improvement.
In summary, our contributions are:
\begin{itemize}    
\item \textbf{An Olympiad-level physics problem benchmark.} \textsc{PhysElite} comprises 11{,}586 Olympiad-tier physics problems, an order of magnitude larger than comparable high-difficulty datasets while maintaining uniform elite difficulty.        
\item \textbf{Multimodal process-level annotations.} Every problem is paired with a visual diagram and a human-verified step-by-step derivation, providing a resource suitable for both rigorous evaluation and process-supervised training.        
\item \textbf{Extensive evaluation and diagnostic analysis.} We benchmark representative open- and closed-source MLLMs on \textsc{PhysElite}, quantify the difference to human experts, compare System-2 and standard models across difficulty levels, and characterize six recurring failure modes that suggest directions for future work.
\end{itemize}

\section{Related Work}







\subsection{Multimodal Scientific Benchmarks}
Early multimodal evaluation suites such as ScienceQA~\citep{lu2022learn}, MathVista~\citep{lu2024mathvista}, MMMU~\citep{yue2024mmmu}, and SciBench~\citep{wang2024scibench} broadened multimodal coverage across disciplines and established the protocol of pairing diagrams with multiple-choice or short-answer items. Their physics subsets, however, are dominated by introductory-level questions and rarely exceed a few hundred items per domain, leaving little headroom for differentiating frontier models. The cross-disciplinary framing further dilutes the physics-specific signal: a model can succeed through generic visual reasoning without engaging the underlying physical laws.

\subsection{Physics-Specific Benchmarks: Scale, Modality, and Difficulty}
Physics-focused resources have grown along three partially competing axes—scale, modality, and difficulty—and to date no single benchmark has satisfied all three.
\begin{table}[t]
\centering
\caption{Comparison of \textbf{\textsc{PhysElite}} with existing physics
reasoning benchmarks. \textsc{PhysElite} is the largest Olympiad-level physics benchmark
with bilingual support, detailed process-level annotations.}
\label{tab:dataset_comparison}
  \begin{minipage}{\textwidth}
    \small
    \textbf{Diagrams} \& \textbf{Solution Step} \& \textbf{Analysis Process}: \cmark = Supported;
    \xmark = Unsupported.
  \end{minipage}
\resizebox{\textwidth}{!}{%
  \begin{tabular}{lrlcccl}
  \toprule
  \textbf{Dataset} & \textbf{Year} & \textbf{Number} & \textbf{Diagrams} & \textbf{Solution Step} &\textbf{Analysis Process} & \textbf{Language} \\
  \midrule
\multicolumn{7}{c}{\textbf{K-12 level}}\\
\midrule
SciBench~\citep{wang2024scibench}    & 2024 & 295         & \cmark & \xmark &\xmark & EN        \\
MMMU~\citep{yue2024mmmu}             & 2024 & 983         & \cmark & \xmark &\xmark  & EN         \\
PhysicsArena~\citep{dai2025physicsarena} & 2025 & 5,103    & \cmark & \xmark &\xmark & EN     \\
PhysUniBench~\citep{physunibench2025}& 2025 & 3,304        & \cmark & \cmark(text only) &\xmark & EN/CN  \\
PHYSICS~\citep{zheng2026scaling} & 2025 & 5,334  & \xmark & \xmark &\xmark & EN/CN    \\
\midrule
\multicolumn{7}{c}{\textbf{College-level}}\\
\midrule
PHYSICS~\citep{zheng2026scaling} & 2025 & 2,127  & \xmark & \xmark &\xmark & EN/CN    \\
UGPhysics~\citep{xu2025ugphysics}    & 2025 & 5,520      & \xmark & \xmark &\xmark & EN/CN  \\
PHYSICS~\citep{feng-etal-2025-physics}    & 2025 & 1,297   & \cmark & \xmark &\xmark & EN    \\
P1-VL (Not Public)~\citep{chen2025p1} & 2026 & 3,907       & \cmark & \xmark &\xmark & EN     \\
\midrule
\multicolumn{7}{c}{\textbf{Olympiad-level}}\\
\midrule
OlympiadBench~\citep{he-etal-2024-olympiadbench} & 2024 & 2,428  & \cmark & \cmark(text only) & \xmark & EN/CN  \\
PHYSICS~\citep{zheng2026scaling} & 2025 & 823  & \xmark & \xmark &\xmark & EN/CN    \\
PHYBench~\citep{qiu2025phybench}     & 2025 & 500          & \xmark & \cmark(text only) &\xmark & EN      \\
PhysReason~\citep{zhang2025physreason} & 2025 & 1,200    & \cmark & \cmark(text only) &\xmark & EN    \\
HiPhO~\citep{yu2025hipho}            & 2025 & 360       & \cmark & \cmark(text only) &\xmark & EN/CN  \\
P1-VL (Not Public)~\citep{chen2025p1} & 2026 & 4,126      & \cmark & \xmark &\xmark & EN     \\

\midrule
\textbf{\textsc{PhysElite} (Ours)} & \textbf{2026} & \textbf{11,586}
  & \textbf{\cmark} & \textbf{\cmark}(Mutimodal)
 &\textbf{\cmark}  & \textbf{EN/CN} \\
\bottomrule
\end{tabular}%
}
\end{table}
\textbf{Scaling along the text-only axis.} PHYSICS~\citep{zheng2026scaling} (8{,}284 bilingual problems) and UGPhysics~\citep{xu2025ugphysics} (5{,}520 problems) provide large training-friendly corpora that span high-school to graduate physics, but both deliberately exclude diagrams to ease ingestion, and neither targets Olympiad difficulty. PHYBench~\citep{qiu2025phybench} curates only 500 problems but pushes them to Olympiad level, illustrating the inverse relationship typical of text-only physics benchmarks: scale is bought at the cost of difficulty, or vice versa.

\textbf{Adding multimodality at undergraduate level.} Recognizing that diagrams, circuits, and ray traces carry information that pure text cannot recover, recent work introduces multimodal physics benchmarks. PhyX~\citep{shen2025phyx} (3{,}000 problems, six reasoning types), PhysUniBench~\citep{physunibench2025} (3{,}304 problems with paired diagrams), PHYSICS~\citep{feng-etal-2025-physics} (1{,}297 problems), and PhysicsArena~\citep{dai2025physicsarena} (5{,}103 high-school CEE-style problems) all couple text with visual input. These benchmarks advance multimodal evaluation but cap difficulty at undergraduate or pre-college level, leaving the Olympiad regime untested. Most are also English-only, limiting cross-lingual analysis.

\textbf{Olympiad-focused benchmarks at limited scale.} Dedicated Olympiad benchmarks probe the upper bound of model capability but are constrained in either scale, modality, or contamination risk. OlympiadBench~\citep{he-etal-2024-olympiadbench} (2{,}428 physics problems) and PhysReason~\citep{zhang2025physreason} (1{,}200 problems, 81\% with diagrams) provide step-level annotations on competition material, yet draw entirely from publicly archived contests, which raises pre-training overlap concerns. HiPhO~\citep{yu2025hipho} compiles 360 problems from 13 recent (2024--2025) Olympiad papers in bilingual form, but its scale precludes per-difficulty statistical analysis. PhoPile~\citep{zheng2026benchmarkingfoundationmodelsretrievalaugmented} expands to 3{,}052 Olympiad problems but remains unimodal and monolingual. Beyond benchmarks, recent work explores reinforcement learning on text-only physics Olympiads~\citep{chen2025p1}, complementary to our multimodal evaluation focus.


\section{\textsc{PhysElite} Dataset}

\subsection{Overview of \textsc{PhysElite}}

\begin{figure}[h]
\centering
\begin{minipage}{0.5\linewidth}
\centering
\footnotesize
\begin{tabular}{lr}
\toprule
\textbf{Statistic} & \textbf{Number} \\
\midrule
Total questions & 11,586 \\
\quad - Mechanics questions & 6,739 \\
\quad - Electronmagenetism questions & 2,811 \\
\quad - Optics questions & 824 \\
\quad - Thermodynamics questions & 652 \\
\quad - Modern Physics questions & 560 \\
\midrule
Difficulties \\
\quad - Easy (1-3) & 32\% \\
\quad - Medium (4-5) & 54\% \\
\quad - Hard (6-7) & 14\% \\
\midrule
Goal Type \\
\quad - Symbolic Expressions & 7,767 \\
\quad - Numerical Values & 1,449 \\
\quad - Qualitative Conclusions & 1,331 \\
\quad - Equations & 1,039 \\
\midrule
Diagrams & 16,130 \\
\midrule
Maximum question length & 2,619 \\
Maximum answer length & 280 \\
Average question length & 135.9 \\
Average answer length & 28.0 \\
\bottomrule
\end{tabular}
\captionof{table}{Key Statistics of \textsc{PhysElite}}
\label{tab:dataset_stats}
\end{minipage}%
\hfill
\begin{minipage}{0.48\linewidth}
\centering
\includegraphics[width=0.85\linewidth]{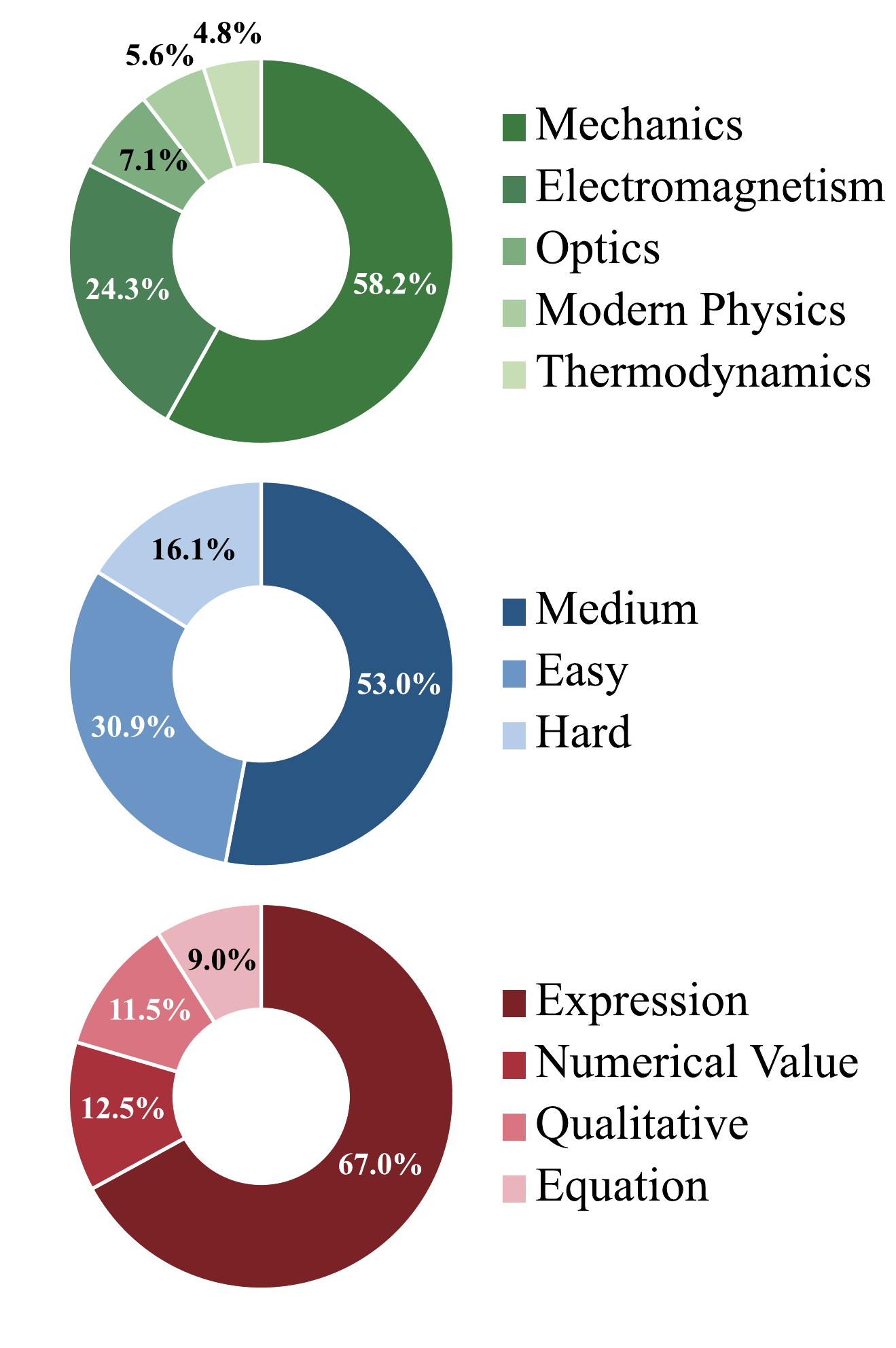}
\captionof{figure}{Distribution of \textsc{PhysElite}}
\label{fig:sankey}
\end{minipage}
\end{figure}

PHYSELITE is, to our knowledge, the first benchmark to combine four desiderata in a single corpus: (i) Olympiad-level difficulty across every problem, (ii) large scale (11,586 problems, an order of magnitude above PHYBench~\citep{qiu2025phybench} and HiPhO~\citep{yu2025hipho}), (iii) bilingual multimodal coverage with paired diagrams and step-level derivations.

Concretely, PHYSELITE is a large-scale, bilingual (Chinese–English) multimodal benchmark for Olympiad-level physics reasoning. It contains 11,586 open-ended problems spanning Mechanics, Electromagnetism, Thermodynamics, Optics, and Modern Physics. Each problem is paired with one or more diagrams, a human-verified step-by-step derivation, the final answer, and metadata including topic tags and a difficulty rating on a 1–7 scale. All problems are sourced from the daily training and practice materials of 15 first-prize-winning students from local secondary schools. This sourcing choice is central to our design: it provides realistic Olympiad-level problems while substantially reducing the pre-training overlap that confounds existing physics benchmarks. More details in Appendix~\ref{app:dataset}.

\subsection{Dataset Format}

An example is presented on the right side of Figure~\ref{fig:annotation}. For each question, we give 1) the question text in both Chinese and English, 2) the schematic diagram and selected process diagram, 3) a detailed, step-by-step solution in natural language, 4) a difficulty level annotated by human experts and 5) its subject category.


To construct the dataset, we first collect approximately 15,000 open-ended problems from the daily learning and practice materials of physics contestants (including textbooks, past competition papers, and training handouts) in image format. These images are scanned at a resolution of 300 DPI to ensure the clarity of text, schematic diagrams, and mathematical symbols, covering a wide range of difficulty levels and subfields fully aligned with international physics competition standards. Each problem image contains the problem text, schematic diagram, step-by-step solution and the final answer.

Then, we convert these 300 DPI problem images into text using the Tesseract OCR engine (v5.3.0), which is widely used for academic text recognition and supports the accurate identification of mixed text and simple symbols. To ensure the accuracy of the converted text—especially for professional physics terms, complex schematic diagram descriptions, and mathematical expressions—we conduct manual verification and correction on all OCR results: any misrecognized characters, missing content, or distorted symbols are revised by professional physics researchers. For the mathematical expressions in answers and step-by-step solutions, we further convert them into \LaTeX{} format manually after verification to preserve the precision of mathematical symbols and logical structure.

After that, we remove duplicated problems using fuzzy matching algorithms (with a similarity threshold of 0.95) to avoid data redundancy. After the above process is finished, the size of the dataset is reduced from around 15,000 to 11,586.

\subsection{Annotation}
\begin{figure}[h]
    \centering
    \includegraphics[width=1\linewidth]{Annotation_Pipeline_and_Sample.pdf}
    \caption{Pipeline of \textsc{PhysElite} data annotation and a sample problem}
    \label{fig:annotation}
\end{figure}

The \textsc{PhysElite} dataset includes two main types of annotations: fine-grained subject categories and difficulty levels. To ensure the accuracy of annotations, we adopted a human-in-the-loop annotation scheme in which LLMs assist human annotators to improve efficiency. The overall annotation process and a sample problem from \textsc{PhysElite} dataset are illustrated in
Figure~\ref{fig:annotation}. More details can be found in appendix~\ref{app:annotation_guidelines}.

\textbf{Difficulty Annotation.} We adopt a two-tier annotation scheme to ensure both accuracy and coverage. When the source materials provide difficulty labels from the original problem setters, we use them directly. For problems without such annotations, human experts assign a difficulty level on a 7-point scale, ranging from 1 (relatively easy) to 7 (extremely hard), referencing the length of the reference solution and the LLM inference token count as auxiliary quantitative signals. Inconsistent or ambiguous cases are resolved through discussion among human expert annotators.

\textbf{Subject categorization.} We define a unified taxonomy covering the major subfields tested in international physics Olympiads, including mechanics, electromagnetism, thermodynamics, optics and modern physics. For problems drawn from sources with a clear chapter structure, we directly map their chapter labels to our taxonomy. For the remaining problems, we first obtain candidate categories through majority voting over predictions from three advanced MLLMs~\citep{anthropic2025claude4,google2025gemini3,openai2025gpt5}, which are then verified by human experts. Finally, an LLM is used to normalize all labels into the unified taxonomy.

\subsection{Dataset Statistics}

Table~\ref{tab:dataset_stats} and Figure~\ref{fig:sankey} summarize the composition of \textsc{PhysElite}. Mechanics and electromagnetism dominate the problems, reflecting the importance of these two subfields in Olympiad training. Difficulty is concentrated in the medium range, with a long tail of hard problems that probe the upper bound of model capability. Answers are open-ended and take four forms: symbolic expressions, numerical values , qualitative conclusions and equations, preventing models from succeeding through multiple-choice shortcuts. The average problem statement is 135.9 words long, with multi-part Olympiad problems reaching up to 2,619 words.

\section{Experiments}

\subsection{Experimental Setup}
\label{sec:experimental setup}
\textbf{Evaluation Models.}
We evaluate a diverse set of models on \textsc{PhysElite}, spanning both LLMs and MLLMs, including 8 open-source and 10 closed-source models. These models fall into two categories:
12 \textbf{System-1 models}, which follow a fast, single-pass reasoning paradigm, and 6 \textbf{System-2 models}, which adopt a slow, iterative long CoT reasoning style.

\textbf{Evaluation Details.}

To establish a human performance baseline, we recruited 15 first-prize-winning students from local secondary schools across different age groups to independently solve the problems, drawing on their usual training records. For multimodal models, the schematic diagram in each problem is included. Each problem is evaluated separately in Chinese and English, with results averaged across the two languages. Decoding hyperparameters and full prompt templates are provided in Appendix~\ref{app:eval}.

\textbf{Scoring.} We report two complementary metrics. The \textit{Answer Score} counts as correct only if its final answer matches the reference. The \textit{Process Score} is computed in two stages: the judge first decomposes the model's response into several key derivation steps, then grades each step as fully correct (1.0), partially correct (0.5), or incorrect (0.0). We grade against the model's own derivation rather than forcing alignment with the reference solution, since physics problems often admit multiple valid solution paths. 

Each response is independently scored by three LLM judges (GPT-5.2, Claude-Opus-4.6, and Gemini 3-Pro) and the final score takes the mean value. We validate the protocol on 800 model responses graded by human experts, including 400 independently graded by a second expert. Full judge prompts are provided in Appendix~\ref{app:eval}, and agreement rates and errors by subject, answer type, and language are reported in Appendix~\ref{app:human_validation}.

\begin{table}[t]
\centering
\caption{Main results on \textsc{PhysElite}, averaged across Chinese and
English evaluations.
\textbf{Ans.}: answer-level accuracy (\%).
\textbf{Mech.}: Mechanics.
\textbf{E\&M}: Electromagnetism.
\textbf{Mod.}: Modern Physics.
\textbf{Therm.}: Thermodynamics.
\textbf{Opt.}: Optics.
\textbf{Proc.}: process score of step-level evaluation.
Best per column in \textbf{bold}, second best
\underline{underlined}.}
\label{tab:main_results}
\vspace{4pt}
\resizebox{\textwidth}{!}{%
\begin{tabular}{lcccccc|c}
\toprule
\textbf{Model}
  & \textbf{Mech.}
  & \textbf{E\&M}
  & \textbf{Mod.}
  & \textbf{Therm.}
  & \textbf{Opt.}
  & \textbf{Ans.}$\uparrow$
  & \textbf{Proc.}$\uparrow$ \\
\midrule
\multicolumn{8}{l}{\textit{Closed-source}} \\
\addlinespace[2pt]
Grok-4.2~\citep{xai2025grok4}\textsuperscript{\dag}                                          & \textbf{34.0}    & \textbf{26.6}    & \textbf{33.0}    & \textbf{35.4}    & \textbf{25.4}    & \textbf{33.7}    & \underline{47.6} \\
Claude-Opus-4.6~\citep{anthropic2025claude4}                           & \underline{27.2} & \underline{20.9} & 24.7             & \underline{33.1} & \underline{21.3} & \underline{28.1} & \textbf{49.6} \\
o3-mini~\citep{openai2025o3mini}\textsuperscript{\dag$\star$}          & 25.3 & 20.4 & \underline{26.8} & 28.4 & 19.6 & 26.2 & 43.9 \\
GPT-5.2~\citep{openai2025gpt5}\textsuperscript{\dag}                                         & 22.1 & 18.6 & 23.9 & 28.4 & 17.4 & 24.0 & 42.6 \\
Gemini-3-Pro~\citep{google2025gemini3}\textsuperscript{\dag}                        & 21.9 & 15.5 & 20.0 & 23.7 & 16.1 & 21.8 & 41.4 \\
Kimi-K2-Thinking~\citep{kimi2025k2}\textsuperscript{\dag}              & 19.3 & 13.2 & 15.6 & 22.8 & 9.6  & 20.0 & 32.7 \\
Claude-Sonnet-4.5~\citep{anthropic2025claude4}                         & 17.5 & 14.7 & 18.5 & 23.7 & 12.6 & 18.5 & 33.6 \\
Gemini-2.5~\citep{geminiteam2025gemini25}\textsuperscript{\dag}                              & 16.9 & 14.2 & 20.7 & 20.9 & 13.5 & 17.9 & 36.8 \\
Qwen-VL-Max~\citep{bai2023qwenvl}                                      & 14.6 & 8.5  & 12.0 & 16.5 & 9.6  & 14.3 & 30.7 \\
GPT-4o~\citep{openai2024gpt4ocard}                                     & 8.8  & 6.5  & 10.2 & 13.8 & 10.0 & 10.4 & 25.0 \\
\midrule
\multicolumn{8}{l}{\textit{Open-source}} \\
\addlinespace[2pt]
Qwen3-VL-235B-A22B~\citep{qwen2025qwen3vl}                             & 19.3 & 15.8 & 19.2 & 22.0 & 14.4 & 20.0 & 36.8 \\
DeepSeek-V3~\citep{liu2024deepseekv3}\textsuperscript{$\star$}         & 18.7 & 10.3 & 14.5 & 19.3 & 10.9 & 17.0 & 34.6 \\
Qwen3-VL-32B~\citep{qwen2025qwen3vl}                                   & 15.2 & 10.1 & 13.8 & 17.8 & 12.2 & 15.9 & 31.5 \\
Qwen3-VL-8B~\citep{qwen2025qwen3vl}                                    & 12.2 & 5.4  & 10.9 & 12.6 & 6.1  & 11.6 & 23.2 \\
Qwen2.5-VL-72B~\citep{bai2025qwen25vl}                                 & 9.1  & 6.2  & 9.1  & 11.1 & 6.1  & 9.8  & 21.5 \\
Dolphin-Mistral-24B~\citep{cognitivecomputations2025dolphin} & 6.1  & 7.5  & 10.1 & 9.9  & 7.0  & 8.6  & 20.6 \\
LLaMA-3.1-70B~\citep{dubey2024llama3}\textsuperscript{$\star$}         & 4.9  & 6.0  & 10.5 & 7.9  & 6.6  & 7.5  & 20.2 \\
Qwen2.5-VL-7B~\citep{bai2025qwen25vl}                                  & 2.4  & 0.5  & 0.7  & 2.0  & 0.5  & 1.7  & 3.7  \\
\midrule
\textit{Human Baseline}                           & 47.6 & 	42.3 & 54.2 & 51.7 & 57.3 & 48.5 & 65.2 \\
\bottomrule
\end{tabular}%
}
\vspace{2pt}
{\small
\textsuperscript{\dag}Reasoning model with extended thinking mode.
\textsuperscript{$\star$}Text-only model; evaluated without
diagram input.
}
\end{table}

\subsection{Main Results}
\label{sec:main results}

Table~\ref{tab:main_results} summarizes answer-level accuracy
and process scores by sub-discipline, averaged across Chinese
and English.

\textbf{Overall difficulty.}
The results presented in Table~\ref{tab:main_results} highlight the inherent difficulty of \textsc{PhysElite}. The best-performing model, Grok-4.2, achieves 33.7\% answer accuracy, followed by Claude-Opus-4.6 at 28.1\%. Notably, all other models fall below the 30\% threshold, underscoring the substantial challenge that \textsc{PhysElite} reasoning poses even for advanced MLLMs. The gap between proprietary and open-weight systems is large: the strongest open-source model (Qwen3-VL-235B-A22B) reaches only 20.0\%. Almost every model also scores higher in English than Chinese; Qwen3-VL-235B-A22B is the lone exception, plausibly reflecting its Chinese-heavy pre-training.

\textbf{Effect of extended thinking.}
Across the 6 reasoning and 12 standard models in our evaluation, extended thinking delivers a clear but uneven advantage. Among the top six entries on PhysElite, four are reasoning models (Grok-4.2, o3-mini, GPT-5.2, Gemini-3-Pro), and the strongest reasoning model (Grok-4.2, 33.7\%) outperforms the strongest standard model (Claude-Opus-4.6, 28.1\%) by 5.6 points. However, the benefit of extended thinking is not uniform: Claude-Opus-4.6 in its default mode surpasses five of the six reasoning models, including GPT-5.2, Gemini-3-Pro, Kimi-K2-Thinking, and Gemini-2.5. This indicates that a sufficiently strong base model without explicit reasoning chains can match or exceed mid-tier reasoning models, and that extended thinking alone is not a substitute for base capability.

\textbf{Process versus answer scores.}
The two metrics expose different aspects of model behavior. For every model, the average step score is higher than answer accuracy, indicating that models often produce partially valid derivations even when their final answer is incorrect. This gap is most pronounced for weaker systems: Claude-Opus-4.6 attains roughly 2.1 times its answer accuracy on the step metric, while LLaMA-3.1-70B reaches about 3.4 times. Models therefore retain substantial credit for early reasoning steps before failing later in the derivation, which motivates the fine-grained error analysis in Section~\ref{sec:analysis}.

\subsection{Effect of Process Diagrams}
\label{sec:process_diagram}

By default, we provide each model only with the schematic diagram. To examine whether additional guidance helps, we evaluate six representative models under two augmented settings: (i) supplying the process diagram alongside the schematic, and (ii) supplying a textual solution hint alongside the schematic. Both settings are compared against the schematic-only baseline on the same problems.

\begin{table}[h]
\centering
\caption{Effect of extra information on selected models.}
\label{tab:intervention_effects}
\renewcommand{\arraystretch}{1.4}
\setlength{\tabcolsep}{5pt}
\resizebox{\textwidth}{!}{%
\begin{tabular}{lcccccc}
\toprule
\textbf{Metric} & \textbf{GPT-5.2} & \textbf{Opus-4.6} & \textbf{Sonnet-4.5} & \textbf{Gemini-2.5} & \textbf{Qwen3-VL-235B} & \textbf{Qwen3-VL-32B} \\
\midrule
\multicolumn{7}{l}{\textit{Process Diagram}} \\
Answer  & 26.7~\textcolor{ForestGreen}{(+2.7)} & 28.4~\textcolor{ForestGreen}{(+0.3)} & 19.1~\textcolor{ForestGreen}{(+0.6)} & 19.8~\textcolor{ForestGreen}{(+1.9)} & 22.5~\textcolor{ForestGreen}{(+2.5)} & 18.5~\textcolor{ForestGreen}{(+2.6)} \\
Process & 46.0~\textcolor{ForestGreen}{(+3.4)} & 51.3~\textcolor{ForestGreen}{(+1.7)} & 35.9~\textcolor{ForestGreen}{(+2.3)} & 40.6~\textcolor{ForestGreen}{(+3.8)} & 40.9~\textcolor{ForestGreen}{(+4.1)} & 37.0~\textcolor{ForestGreen}{(+5.5)} \\
\midrule
\multicolumn{7}{l}{\textit{Solution Hint}} \\
Answer  & 23.5~\textcolor{red}{(-0.5)} & 28.4~\textcolor{ForestGreen}{(+0.3)} & 18.5~\textcolor{black}{(+0.0)} & 17.9~\textcolor{black}{(+0.0)} & 19.9~\textcolor{red}{(-0.1)} & 15.7~\textcolor{red}{(-0.2)} \\
Process & 42.8~\textcolor{ForestGreen}{(+0.2)} & 50.0~\textcolor{ForestGreen}{(+0.4)} & 37.0~\textcolor{ForestGreen}{(+2.3)} & 36.7~\textcolor{red}{(-0.1)} & 36.8~\textcolor{black}{(+0.0)} & 32.4~\textcolor{ForestGreen}{(+0.9)} \\
\bottomrule
\end{tabular}%
}
\end{table}

Table~\ref{tab:intervention_effects} reports the results. Adding process diagrams improves both metrics for all six models, with process score gains consistently larger than answer score gains. In contrast, providing textual solution hints yields only marginal changes, and in several cases even degrades answer accuracy. This contrast suggests that visual process diagrams provide substantially stronger guidance for intermediate reasoning than equivalent textual hints, highlighting the unique value of diagrammatic information.

\section{Analysis}
\label{sec:analysis}

We conduct three analyses to characterize \textbf{where}, \textbf{how}, and on \textbf{which} problems modern multimodal LLMs fail: (1) Step-position localization identifies where the first error appears along the derivation chain. (2) Failure-severity decomposition distinguishes execution slips from fundamental reasoning errors. (3) Problem-class stratification examines failure patterns across sub-disciplines and difficulty levels.

\subsection{Where the first error appears.}

We define the first-error index as the 1-based position of the earliest step graded below $1.0$ in the model's derivation. Stronger models reach deeper into the derivation before failing: Claude-Opus-4.6 makes its first error at step $2.48$, while weaker open-source models fail near step $1.4$, with Qwen2.5-VL-7B failing at step $1$ in $96.8\%$ of cases. GPT-4o ($1.57$) and Kimi-K2-Thinking ($1.62$) sit closer to small open-source models, suggesting that first-error depth tracks base capability rather than vendor or training paradigm. However, depth alone does not determine outcome. Grok-4.2 fully completes $21.8\%$ of problems, roughly $30\%$ more than Claude-Opus-4.6 at similar first-error depth.

\subsection{Failure-severity decomposition}

\textbf{Soft versus hard failures.}
For each failed step, we ask whether the judge assigned partial correct or totally wrong. The fraction of soft failures ranges from $7\%$ for Qwen2.5-VL-7B to $46\%$ for Claude-Opus-4.6, with the four highest values held by closed-source frontier models. This fraction is consistently higher in Chinese than in English across all $18$ models, with gaps up to $12$ points.

\textbf{Last-step-only failures.}
Some failures occur only at the final step, with all preceding steps fully correct. These failures peak at $7.8\%$ for Claude-Opus-4.6. The pattern tracks capability, as weaker models rarely reach the closing step. The errors involved are typically numerical substitution, sign handling, or final-form simplification, all of which leave the upstream reasoning intact.

\subsection{Problem-class stratification}

\textbf{Optics is the universal weak point.}
First-step error rates exhibit a stable cross-model topical ordering: setup failures are most frequent in Optics and least frequent in Thermodynamics, with Mechanics, Electromagnetism, and Modern Physics in between. Among the top eight closed-source models, the Optics$-$Thermodynamics gap exceeds $20$ percentage points on average, and the same ordering holds for weaker models. The Optics gap suggests that current MLLMs handle geometric and phase-based reasoning less reliably than the algebraic manipulation required by other sub-disciplines.

\textbf{Hard problems are not the worst.}
Most models show a U-shape with Medium as the lowest-scoring slice, with the Hard-Medium gap exceeding $10$ points for four frontier reasoning models (Gemini-2.5, Claude-Sonnet-4.5, GPT-5.2, Claude-Opus-4.6). Only Grok-4.2, Kimi-K2-Thinking, and Qwen2.5-VL-7B show the expected monotone decrease from Easy to Hard. Human contestants follow the monotone trend, indicating that the inversion reflects training rather than intrinsic difficulty. We attribute the U-shape to targeted post-training on Olympiad competition problems, which lifts Hard performance while leaving the daily-practice Medium subset as a genuinely out-of-distribution test.

\subsection{Error Analysis}

\begin{figure}[h]
    \centering
    \includegraphics[width=1\textwidth]{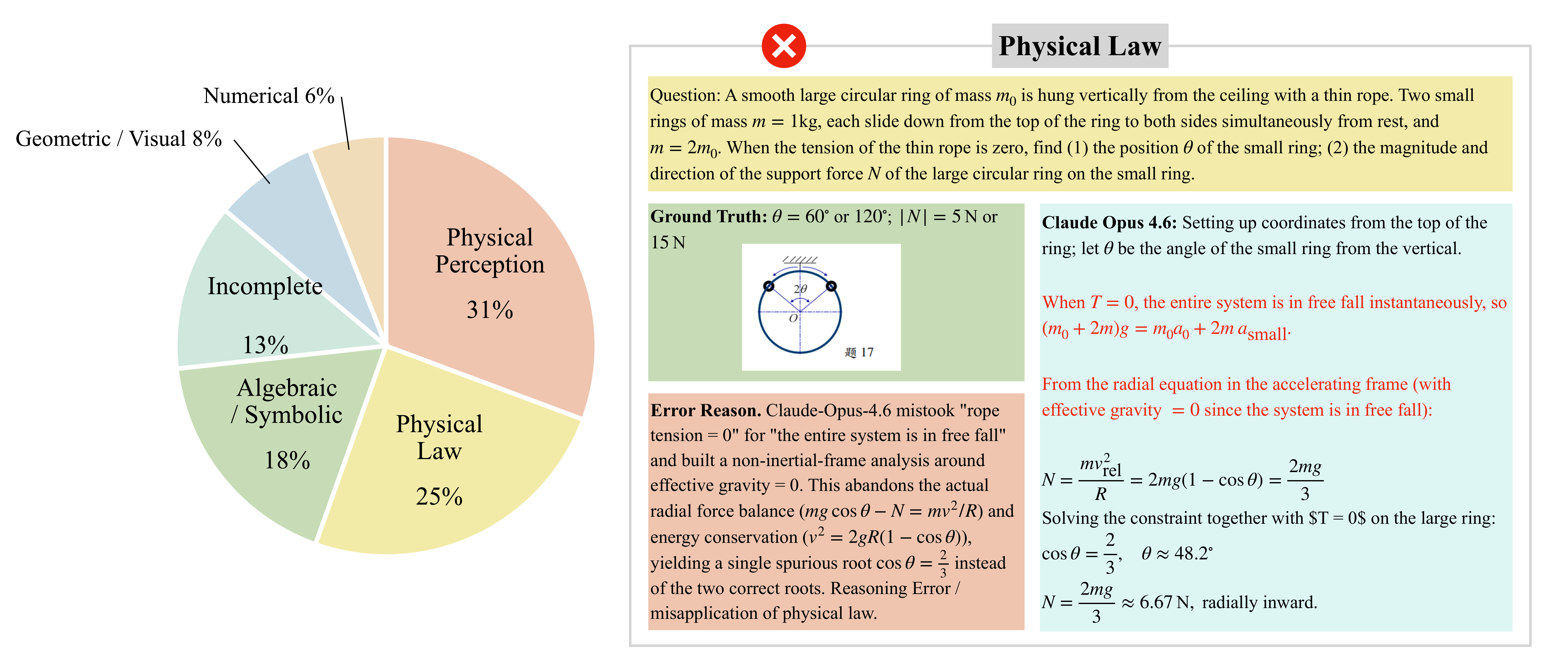}
    \caption{Error distribution of Claude Opus 4.6 and an example of physical law error.}
    \label{fig:erroranalysis}
\end{figure}

We randomly sample $100$ incorrect predictions from Claude-Opus-4.6 and classify each into six error categories. The error taxonomy and proportions are visualized in Figure~\ref{fig:erroranalysis}.  Among all error types, the dominant failures occur at the front end of the reasoning pipeline (problem reading, principle selection, and geometric setup). Physical perception ($31\%$), misapplication of physical laws ($25\%$), and geometric or visual misreadings ($8\%$) together account for roughly two thirds of all failures, indicating that the model more often fails to set up the problem than to execute the subsequent algebra. Algebraic and symbolic mistakes ($18\%$) and incomplete derivations ($13\%$) make up most of the rest, while purely numerical errors are a distant last at $6\%$. An illustrative physical law error is shown on the right side of Figure~\ref{fig:erroranalysis}, where the model incorrectly reads ``rope tension is zero'' as implying free fall of the entire system, switches to a non-inertial analysis, and misses one of the two correct solution branches. More case studies can be found in the appendix~\ref{app:error}.

\section{Conclusion}

We presented \textsc{PhysElite}, a large-scale bilingual multimodal benchmark for Olympiad-level physics reasoning, and showed that current frontier LLMs remain far from expert-level performance under both answer-level and process-level metrics. Beyond reporting aggregate accuracy, our step-level protocol reveals where reasoning chains fail and which error types dominate across topics and model families. Our analyses further indicate that a single nominal difficulty score is insufficient for cross-benchmark interpretation: source distribution and problem style alignment can dominate model outcomes, especially when compared with competition-style sets such as iPhO-like evaluations. We hope \textsc{PhysElite} can serve both as a rigorous evaluation benchmark and as a process-supervised resource for training stronger scientific reasoners.

\section{Limitations and future works}
Our benchmark has a few limitations: partial-credit scoring introduces minor annotation ambiguity in borderline cases, and our training data may have minor overlap with commercial model datasets. We aim to mitigate these issues in future work, particularly by collecting a fully private dataset to reduce potential data leakage risks. We hope \textsc{PhysElite} can serve both as a rigorous evaluation benchmark and as a process-supervised resource for training stronger scientific reasoners.

The human baseline reflects the performance of students with extensive Olympiad training. Familiarity with common problem styles, physical models, and solution methods can influence performance on new questions. The human evaluation protocol and this familiarity effect are described in Appendix~\ref{app:human}.




\begin{ack}
This work was supported by National Natural Science Foundation of China under No. 62436009, 62276258 and Jiangsu Science and Technology Programme BK20251812, and Open Research Fund of The State Key Laboratory of Multimodal Artificial Intelligence Systems.
\end{ack}

\bibliographystyle{unsrtnat} 
\small
\bibliography{Reference}  
\normalsize


\newpage

\appendix
\section{More Detailed Statistics about \textsc{PhysElite}}

\label{app:dataset}

This appendix provides additional details about \textsc{PhysElite}, following the same order as the main paper: dataset composition, construction and annotation, evaluation protocol, scoring, and extended analysis. We use the appendix to make the benchmark easier to inspect and reproduce, while keeping the main text focused on the central findings.

\subsection{Distribution of Text Length}

\begin{figure}[h]
    \centering
    \includegraphics[width=1\linewidth]{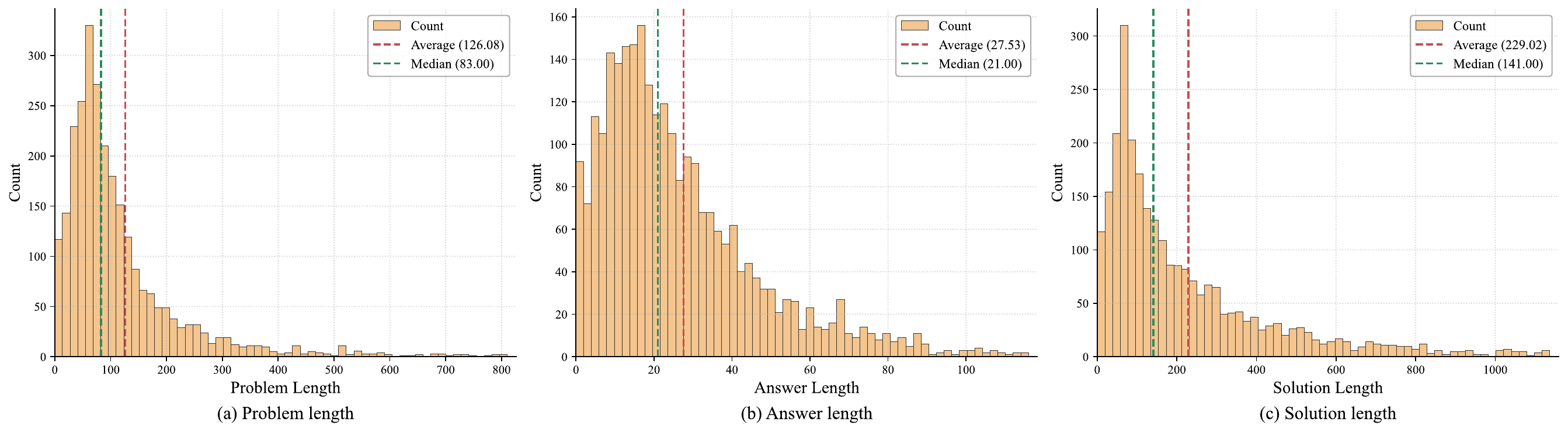}
    \caption{The distribution of the length per problem in \textsc{PhysElite}.}
    \label{fig:phys_length_distribution}
\end{figure}

Questions in \textsc{PhysElite} are presented in English or Chinese. As shown in Table~\ref{tab:dataset_stats}, the longest question in \textsc{PhysElite} spans 2,619 words, with an average length of 280 words. Figure~\ref{fig:phys_length_distribution} further illustrates the distribution of text lengths, highlighting the diversity of \textsc{PhysElite}. The length of English and Chinese questions is counted in words and Chinese characters, respectively.

\subsection{Difficulty and Answer Types}

Each problem is assigned a difficulty score on a 1--7 scale. In the main experiments, we group these labels into three ranges: Easy (1--3), Medium (4--5), and Hard (6--7). The resulting dataset is concentrated around the medium range, with a long hard tail. This is expected for elite daily-practice materials: many problems are designed to develop reusable reasoning skills rather than simply reproduce the hardest official contest items.

\begin{table}[h]
\centering
\caption{Difficulty and answer-type distribution of \textsc{PhysElite}.}
\label{tab:app_difficulty_answer}
\begin{tabular}{lrr}
\toprule
\textbf{Type} & \textbf{Count / Share} & \textbf{Description} \\
\midrule
Easy (1--3) & 32\% & Shorter or more direct derivations \\
Medium (4--5) & 54\% & Multi-step Olympiad-style reasoning \\
Hard (6--7) & 14\% & Long derivations or advanced modeling \\
\midrule
Symbolic expressions & 7,767 & Closed-form symbolic answers \\
Numerical values & 1,449 & Numeric answers with units or constants \\
Qualitative conclusions & 1,331 & Physical judgments or comparisons \\
Equations & 1,039 & Relations, constraints, or derived equations \\
\bottomrule
\end{tabular}
\end{table}

The open-ended answer format is a central part of the benchmark design. Unlike multiple-choice settings, models must produce the final physical quantity or relation directly. This makes answer-level scoring stricter and gives the process score a larger diagnostic role.

\subsection{Distribution of Image Number per question}

\begin{figure}
    \centering
    \includegraphics[width=0.8\linewidth]{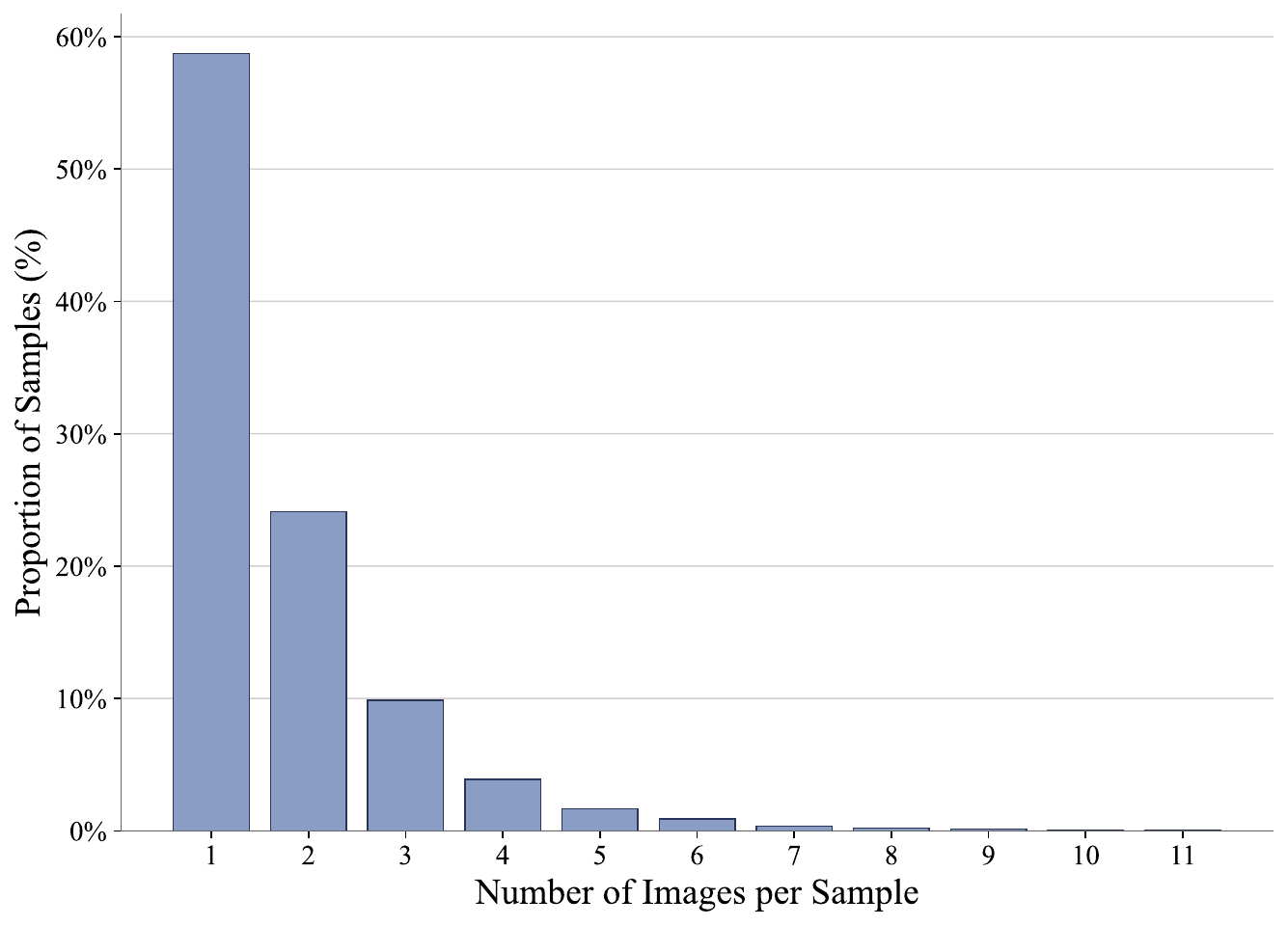}
    \caption{The distribution of the number of images per question in \textsc{PhysElite}.}
    \label{fig:phys_image_count_distribution}
\end{figure}

As shown in Figure~\ref{fig:phys_image_count_distribution}, the majority of questions in the \textsc{PhysElite} dataset (58.7\%) are accompanied by a single image. The remaining questions are associated with multiple images, with the number of images ranging from two to eleven.

\section{Introduction of our Fine-grained Physics Subjects}
\label{app:taxonomy}
\begin{figure}[h]
    \centering
    \includegraphics[width=1\linewidth]{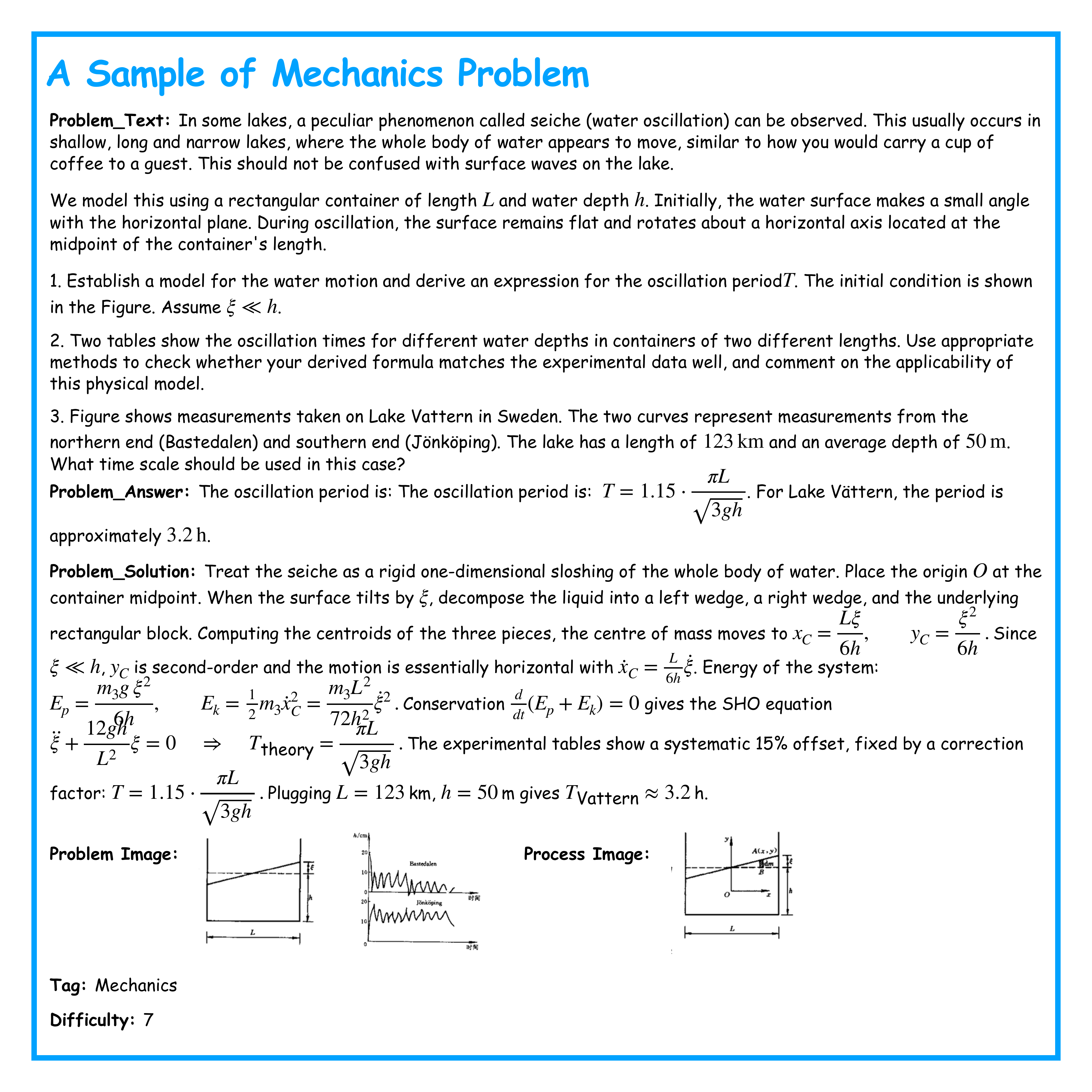}
    \caption{A sample of Mechanics Problem}
    \label{fig:mechanics}
\end{figure}

\begin{figure}[h]
    \centering
    \includegraphics[width=1\linewidth]{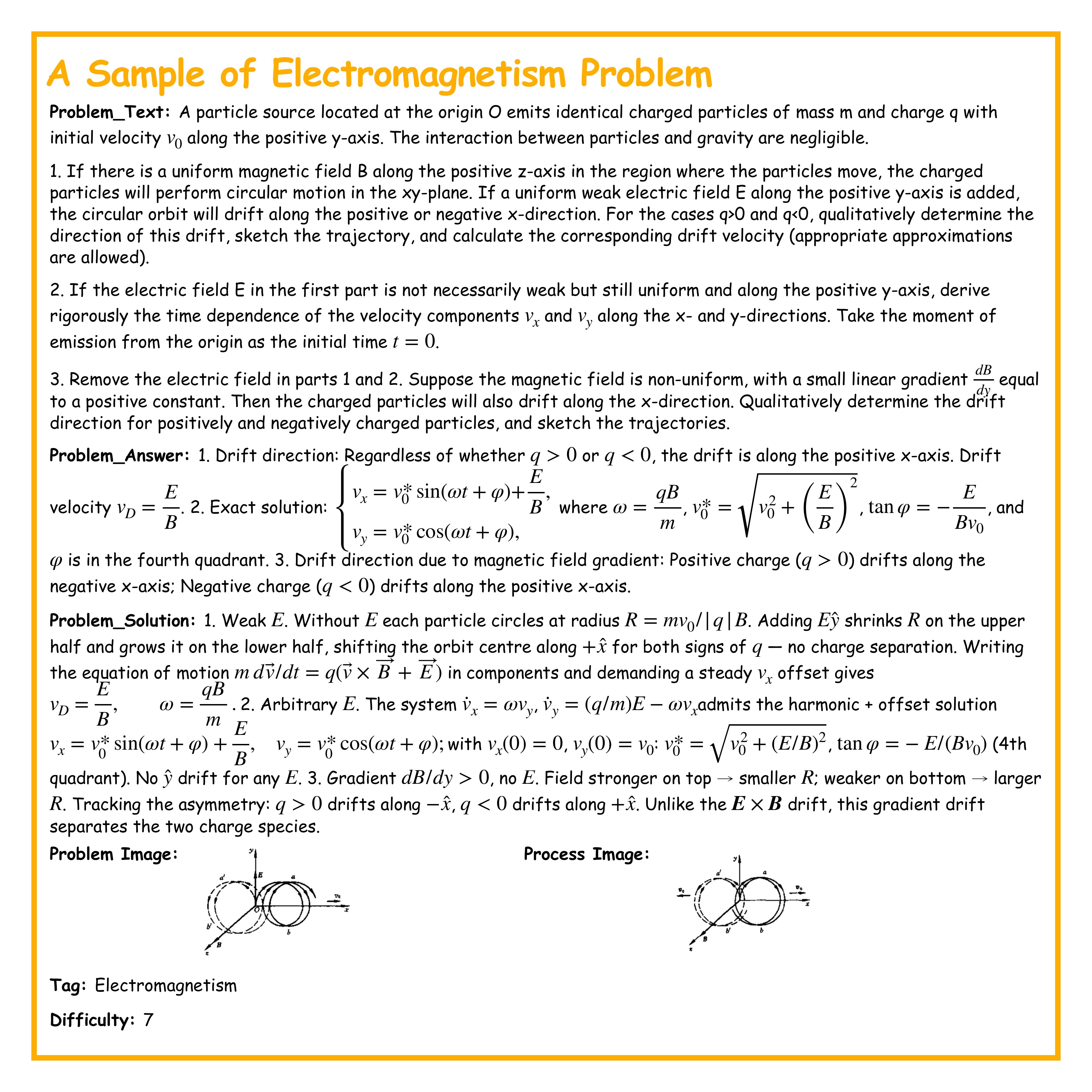}
    \caption{A sample of Electromagnetism Problem}
    \label{fig:electromagnetism}
\end{figure}

\begin{figure}
    \centering
    \includegraphics[width=1\linewidth]{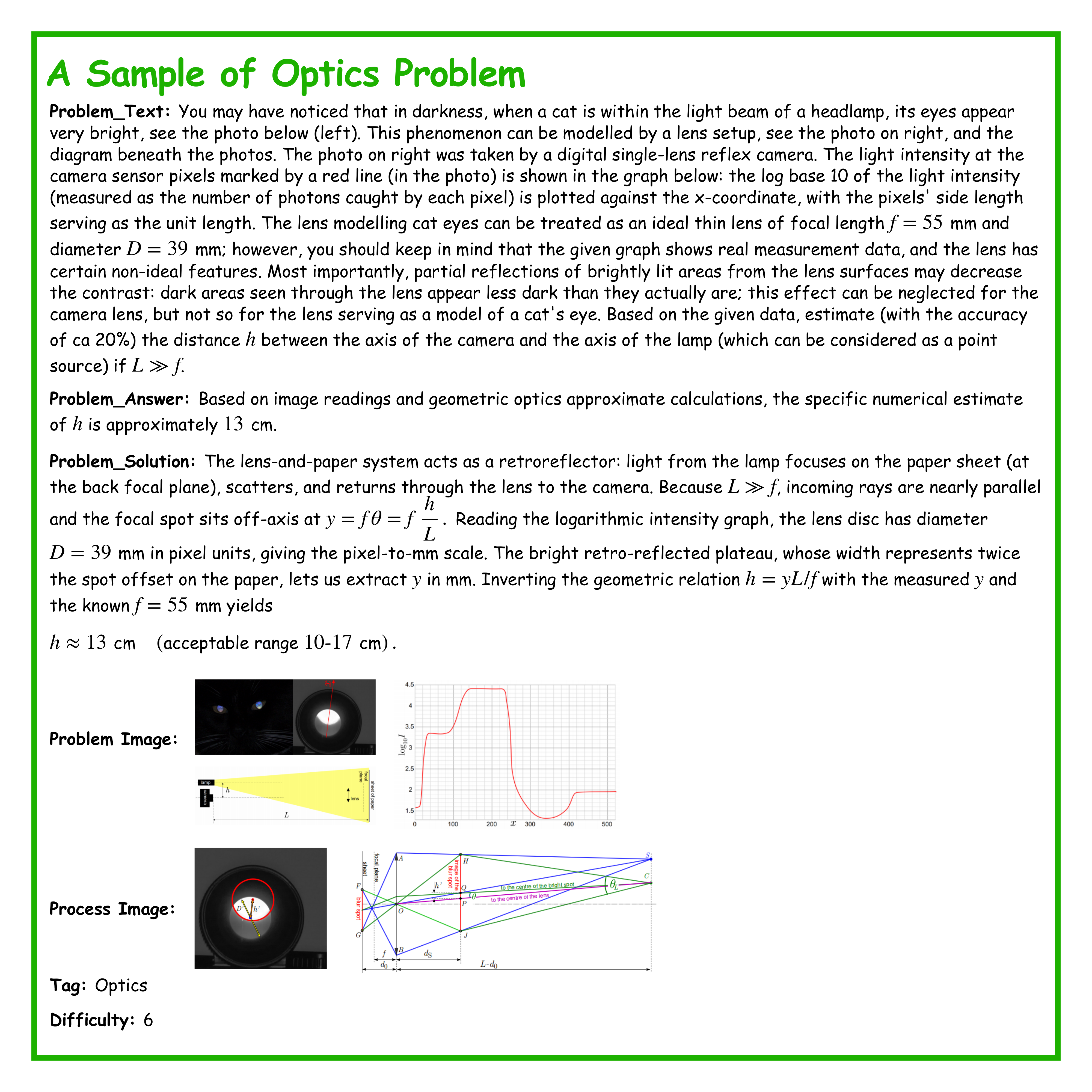}
    \caption{A sample of Optics Problem}
    \label{fig:optics}
\end{figure}

\begin{figure}
    \centering
    \includegraphics[width=1\linewidth]{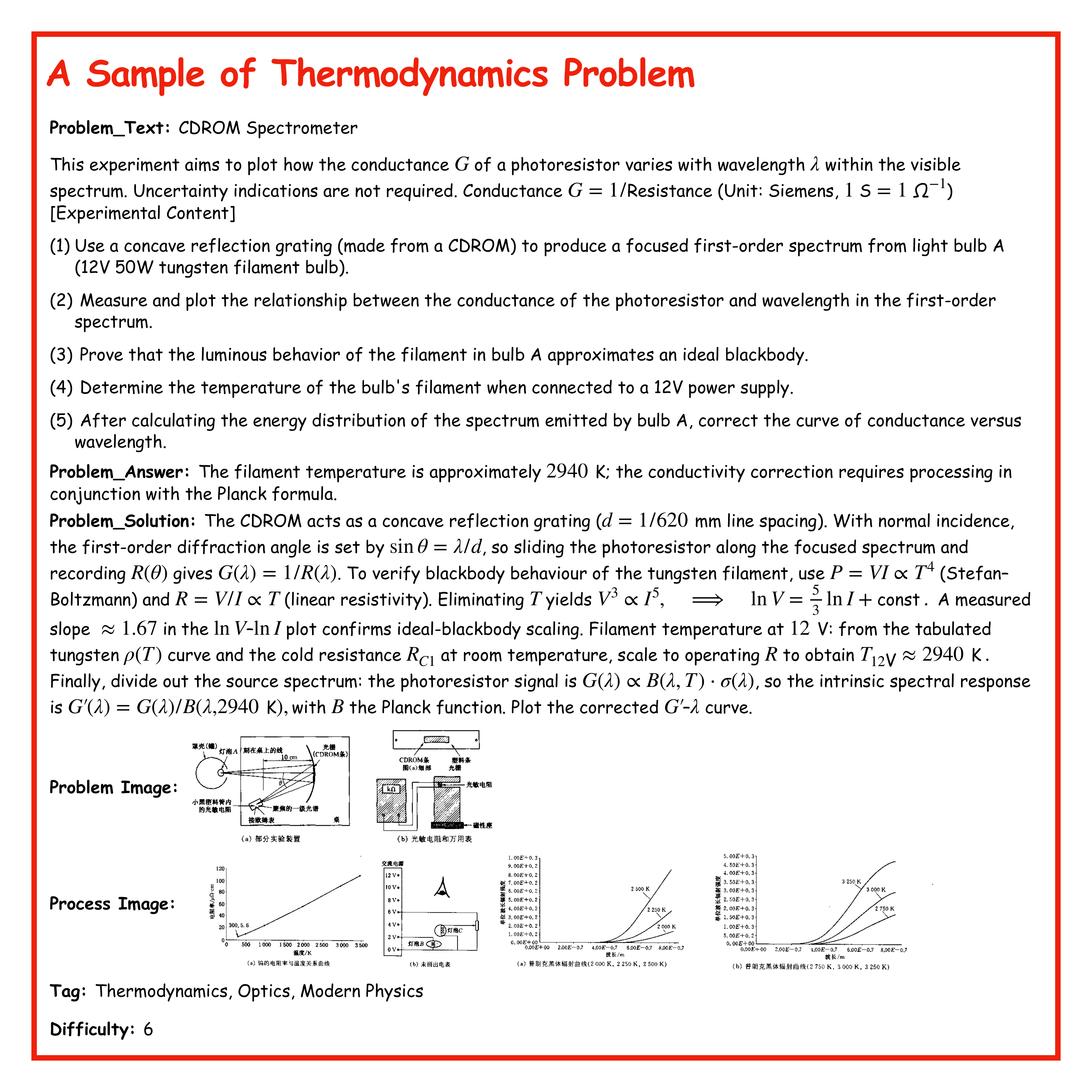}
    \caption{A sample of Thermodynamics Problem}
    \label{fig:thermodynamics}
\end{figure}

\begin{figure}
    \centering
    \includegraphics[width=1\linewidth]{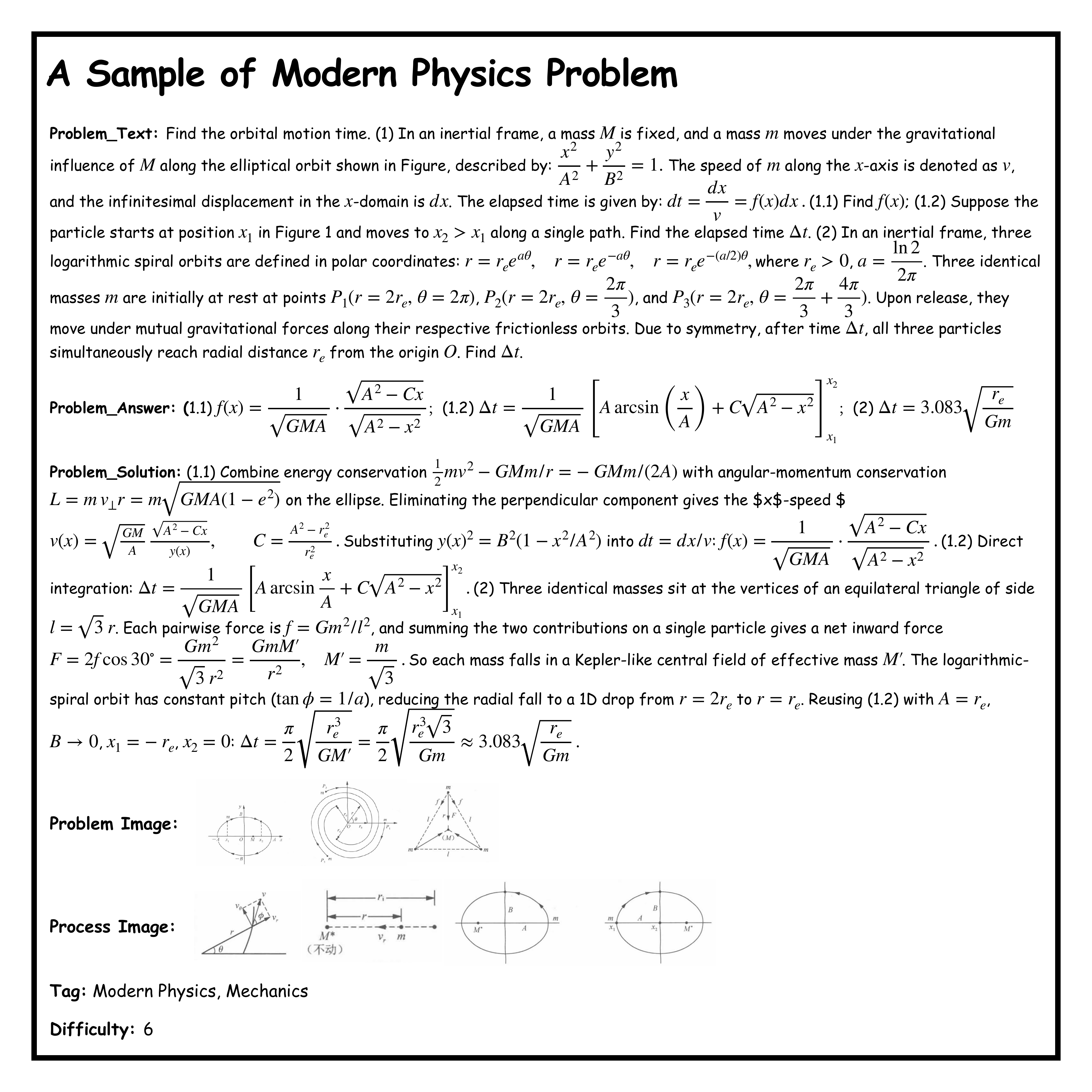}
    \caption{A sample of Modern Physics Problem}
    \label{fig:modern}
\end{figure}

In this section, we introduce the five physics subject categories defined in our \textsc{PhysElite} benchmark. These categories are designed to reflect the diverse reasoning skills required for Olympiad-level physical problem solving. Rather than treating physics as a monolithic domain, we present a structured taxonomy that decomposes the problem space into conceptually distinct yet complementary components. Each problem is assigned one or more subject labels according to the main conceptual obstacle in its reference solution. The taxonomy is intentionally broad rather than overly fine-grained, because many Olympiad problems combine multiple physical principles, and a coarser partitioning yields more interpretable performance diagnostics. Figure~\ref{fig:mechanics} to Figure~\ref{fig:modern} illustrate representative image examples
corresponding to each category.

Our classification is based on both the underlying physical principles (e.g., conservation laws, field theory, statistical reasoning) and the characteristic reasoning patterns elicited by each domain (e.g., model construction, equivalent-circuit reduction, ray geometry analysis). This allows for more interpretable performance diagnostics and supports targeted model evaluation. The five categories span a broad range of physics competencies, ensuring coverage of foundational classical topics as well as advanced modern physics skills.

\textbf{1. Mechanics.} This category focuses on the analysis of motion, forces, and energy in classical systems, encompassing kinematics, dynamics, rigid-body motion, oscillations, waves, gravitation, fluids, and energy--momentum reasoning. Problems typically require constructing a physical model from a diagram, choosing appropriate coordinate systems, and combining force balance, conservation laws, and geometric constraints. These problems often serve as foundational tests of physical reasoning and demand the integration of mathematical manipulation with physical intuition. \emph{Example task:} Determine the maximum compression of a spring when a block slides down an inclined plane and collides with it, given the friction coefficient and initial height.

\textbf{2. Electromagnetism.} This subject class targets reasoning about electric and magnetic phenomena, including electrostatics, direct- and alternating-current circuits, magnetic fields, electromagnetic induction, and charged-particle motion. The focus lies in field-level reasoning and the analysis of dynamical responses to electromagnetic interactions. Typical reasoning patterns include field superposition, equivalent-circuit reduction, Lorentz-force dynamics, and time-varying flux analysis. \emph{Example task:} Compute the induced EMF in a rotating conducting rod within a non-uniform magnetic field, or determine the trajectory of a charged particle in crossed electric and magnetic fields.

\textbf{3. Thermodynamics.} This category involves reasoning about heat, work, and the macroscopic behavior of thermal systems, particularly those that rely on state variables and process constraints. It covers heat transfer, ideal-gas processes, phase changes, entropy-related reasoning, and cyclic processes. Solutions typically require careful interpretation of state variables and process paths rather than purely algebraic manipulation, and demand an understanding of how thermodynamic constraints couple with mechanical or chemical configurations. \emph{Example task:} Analyze the efficiency of a non-standard thermodynamic cycle on a $P$--$V$ diagram, or determine the equilibrium temperature when two gases at different states are connected through a valve.

\textbf{4. Optics.} This class includes problems involving the propagation, interaction, and interference of light, emphasizing the ability to switch between geometric and wave-based descriptions. It spans geometric optics, lenses and mirrors, interference, diffraction, polarization, and optical-path reasoning. Solvers must mentally trace rays through optical systems or track phase relations across multiple paths. As discussed in the main analysis, optics is a persistent weak point for current models, partly because the setup often depends on precise ray geometry or phase relations that are difficult to extract from text and diagrams alone. \emph{Example task:} Determine the position and magnification of the final image formed by a two-lens system, or compute the fringe spacing in a modified double-slit experiment with an inserted glass plate.

\textbf{5. Modern Physics.} This is the most specialized category, featuring problems that go beyond classical mechanics and electromagnetism, including special relativity, quantum phenomena, atomic physics, and nuclear physics. It involves applying principles such as Lorentz transformations, energy--level quantization, photon--matter interactions, and nuclear decay laws. Problems in this category often require specialized principles that are less frequently encountered in routine high-school datasets, and bridge classical reasoning with the conceptual frameworks of modern physics. \emph{Example task:} Compute the wavelength shift of a photon undergoing Compton scattering, or determine the relativistic energy of a particle produced in a decay process.

\clearpage

\section{More Detailed Construction of \textsc{PhysElite}}
\label{app:construction}

\subsection{Data Collection Pipeline}

We start from approximately 15,000 open-ended problems collected from daily learning and practice materials of physics contestants, together with private exam sets used in advanced training. The source materials are image-based and typically contain the problem statement, one or more diagrams, the step-by-step solution, and the final answer. Compared with official competition-only datasets, these materials provide a broader picture of the problem styles used in sustained Olympiad preparation.

The construction pipeline has four stages. First, source images are screened to retain problems with complete statements, solutions, and answers. Second, the problem text and solution are manually transcribed into structured records, with mathematical expressions converted into \LaTeX{} to preserve symbolic precision. Third, diagrams are extracted and linked to the corresponding problem identifiers. Finally, duplicated or near-duplicated problems are removed through fuzzy matching and human review. After this process, the dataset size is reduced from roughly 15,000 candidates to 11,586 curated problems.

\subsection{Dataset Format}

\label{app:format}

Each problem is stored as a structured record. The key fields are:

\begin{itemize}
    \item \textbf{\texttt{problem\_id}}: a unique identifier assigned to each problem.
    \item \textbf{\texttt{problem\_text\_cn} / \texttt{problem\_text\_en}}: the Chinese and English versions of the problem statement.
    \item \textbf{\texttt{problem\_solution\_cn} / \texttt{problem\_solution\_en}}: the full step-by-step solution in both languages.
    \item \textbf{\texttt{problem\_answer}}: the final answer used for answer-level scoring.
    \item \textbf{\texttt{problem\_img}}: filenames of schematic diagrams that describe the physical setup.
    \item \textbf{\texttt{problem\_img\_process}}: filenames of process diagrams that visualize intermediate solution steps.
    \item \textbf{\texttt{subject}} and \textbf{\texttt{difficulty}}: the normalized subject label and 1--7 difficulty label.
\end{itemize}
This format is designed to support both direct answer evaluation and process-level analysis. The same record can be used to evaluate text-only models, multimodal models with schematic diagrams, and process-supervised methods that require reference derivations.

\begin{figure}
    \centering
    \includegraphics[width=1\linewidth]{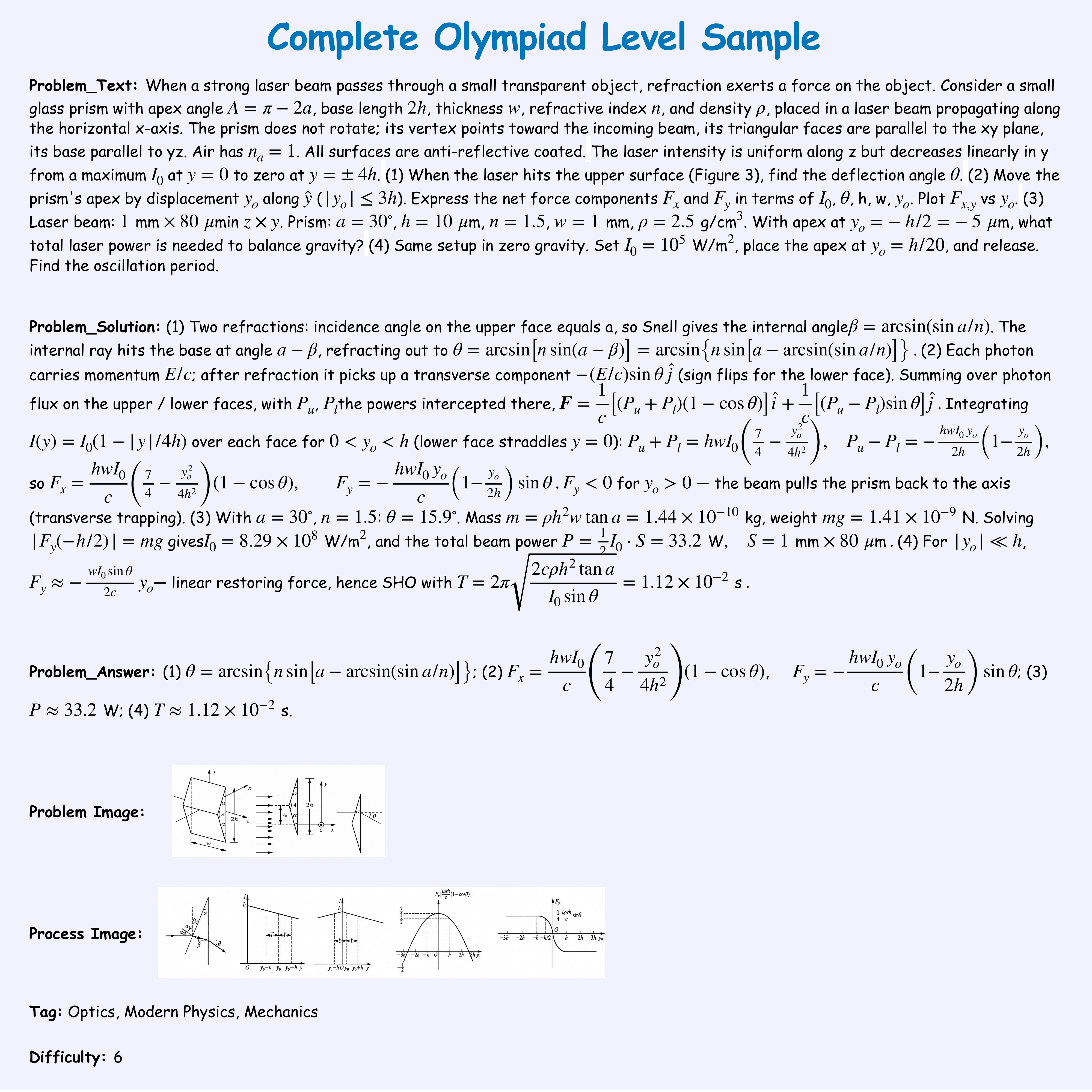}
    \caption{A complete sample problem of our Dataset}
    \label{fig:completesample}
\end{figure}

\subsection{Source Diversity and Representativeness}
Many high-difficulty physics benchmarks rely heavily on official competition problems. Such questions are valuable, but they are designed for specific contests, years, and scoring rubrics. As a result, their topic and reasoning distributions can be narrow. \textsc{PhysElite} instead emphasizes daily contestant training materials and private exam sets. This choice broadens the assessment scenarios: the dataset includes routine elite-practice problems, medium-difficulty problems that teach transferable modeling patterns, and hard problems that approach national or international Olympiad style.
This source design is important for interpreting model behavior. In the main analysis, we observe a U-shaped difficulty trend for many models: medium problems can be harder for models than the nominal hard subset. This suggests that source style and solution-template familiarity influence model performance in addition to intrinsic physics difficulty.
\section{Annotation Protocol}
\label{app:annotation_guidelines}
The annotation process contains three stages: transcription, difficulty annotation, and subject categorization. Human annotators are responsible for final decisions, while LLMs are used only to improve efficiency in candidate generation or label normalization.
\subsection{Transcription Guidelines}

Annotators are asked to act as faithful transcribers. The goal is to convert the source image into structured text without altering the underlying physics content. Mathematical expressions are converted to \LaTeX{}; original symbols, subscripts, and naming conventions are preserved whenever possible. Multi-part problems are kept within a single record, and all sub-answers are included in the final answer field.

If a portion of the source image is unclear, annotators mark the span for review rather than guessing. When source diagrams contain labels or annotations, these are preserved in the image and also reflected in the bilingual diagram description when needed.
\subsection{Difficulty Annotation}
Each problem receives an integer difficulty label from 1 to 7. When the source material provides a difficulty label from the original problem setter, we use it directly. Otherwise, human experts assign the label using the following anchors:

\begin{itemize}
    \item \textbf{1--2}: relatively easy problems solvable in a few short steps.
    \item \textbf{3--4}: introductory Olympiad or routine undergraduate-style problems requiring multiple principles or a non-trivial setup.
    \item \textbf{5--6}: hard Olympiad-style problems requiring careful modeling, multi-step derivation, or coordination of several physical concepts.
    \item \textbf{7}: extremely hard problems comparable to the most demanding national or international Olympiad items.
\end{itemize}
Annotators may adjust the rating by one level when a problem combines several subfields, has an unusually long solution, requires advanced mathematical techniques, or depends on specialized physical knowledge. Ambiguous cases are resolved through discussion among human expert annotators.

\subsection{Subject Categorization}

Each problem is assigned to one of five categories: Mechanics, Electromagnetism, Thermodynamics, Optics, or Modern Physics. If the source has a clear chapter structure, the chapter label is mapped directly to the taxonomy. Otherwise, three advanced MLLMs propose candidate labels, and the majority-vote candidate is reviewed by human experts. The human judgment is final. After all labels are assigned, an LLM-based normalization pass standardizes surface variants such as ``E\&M'' and ``Electromagnetism'' without changing the underlying category assignment.

For cross-domain problems, annotators select the category corresponding to the core reasoning challenge. For example, if a problem combines charged-particle motion with mechanical oscillation, the label depends on whether the decisive step is electromagnetic force modeling or mechanical dynamics.

\subsection{Data Quality Audit}
\label{app:data_quality_audit}

We randomly sampled 800 problems for a data quality audit. We re-annotated these problems using the dataset construction protocol and compared the new annotations with the released records. The audit covered transcription, translation, subject labels, difficulty labels, and solution steps.

We checked transcriptions against the source scans for changes to words, mathematical symbols, formulas, and numerical values, as well as missing content. We checked English translations sentence by sentence against the Chinese source for changes in physical meaning, terminology, and problem conditions. Subject labels were reassigned according to the main reasoning challenge in the reference solution. Difficulty labels were reassigned using the seven-point annotation scale. A difference of two or more levels counted as an error; a one-level difference counted as normal variation between annotators. Reference solutions were checked step by step for missing reasoning, inconsistent steps, and missing formulas. Differences in wording, formatting, and mathematically equivalent solution paths were accepted.

The audit found five subject-label discrepancies and seven difficulty-label errors. Table~\ref{tab:data_quality_audit} summarizes the results.

\begin{table}[htbp]
\centering
\caption{Data quality audit of 800 randomly sampled problems.}
\label{tab:data_quality_audit}
\begin{tabular}{lrr}
\toprule
\textbf{Audit item} & \textbf{Count} & \textbf{Rate} \\
\midrule
Transcription & 0/800 & 0\% \\
Translation & 0/800 & 0\% \\
Subject labels & 5/800 & 0.625\% \\
Difficulty labels & 7/800 & 0.875\% \\
Solution steps & 0/800 & 0\% \\
\bottomrule
\end{tabular}
\end{table}

\section{Annotator and Human-Level Evaluation}
\label{app:human}
Human experts resolve uncertain difficulty labels and subject labels during dataset construction. They also grade model responses to validate the process-scoring protocol, as described in Appendix~\ref{app:human_validation}.

\subsection{Human Baseline Protocol}
\label{app:human_baseline_protocol}

The human baseline combines archived first attempts and additional timed attempts. For problems covered by the participating students' training archives, we collected their recorded first attempts and rescored them using the evaluation pipeline. For problems without archived attempts, an additional group of fifteen students solved them individually under closed-book conditions, with a limit of 45 minutes per problem. Each student was provided with a Casio fx-991CNX calculator.

The students contributing to the human baseline were distinct from those whose training materials were used to construct the dataset. Before the additional timed evaluation, problems were screened individually. Any problem that a student reported having seen was excluded from that student's set.

Each participant contributed responses to about 2,000 problems on average. Each problem had responses from two to three students. We retained the highest score for each problem and averaged these scores across problems. Human responses were scored using the same pipeline as model responses.

We also report a human baseline in Table~\ref{tab:main_results}. Human contestants reach 48.5\% answer accuracy and 65.2 process score overall, with answer accuracy of 47.6\% in Mechanics, 42.3\% in Electromagnetism, 54.2\% in Modern Physics, 51.7\% in Thermodynamics, and 57.3\% in Optics. The gap between this baseline and the best MLLM result shows that \textsc{PhysElite} remains challenging even for frontier models.

\subsection{Familiarity}

The participants had extensive experience with Olympiad training. Familiarity with common problem styles, physical models, and solution methods can influence performance on new questions. The human baseline therefore describes the performance of trained contestants, including the contribution of this experience.

\section{Evaluation Details}
\label{app:eval}
\subsection{Evaluation Settings}

For multimodal models, the schematic diagram is included together with the problem text. For text-only models, only the textual problem statement is provided. Each problem is evaluated separately in Chinese and English, and the reported results average across the two languages.

The evaluation includes 18 models: 10 closed-source models and 8 open-source models. Six of the evaluated models are treated as System-2 reasoning models with extended thinking mode, while the remaining twelve follow a standard System-1 response style. Text-only models are marked in Table~\ref{tab:main_results} and are evaluated without diagram input.

All eighteen models completed the main evaluation without external tools. Code execution, retrieval, and calculators were disabled. Models received the problem statement and, for multimodal models, the statement diagrams. Each model generated its full derivation and final answer within its response. Tool-assisted results are reported separately in Appendix~\ref{app:tool_assisted}.

\subsection{Prompt for Response Generation}
To ensure the model provides accurate responses, we use the following CoT prompt for both Multimodal Large Language Models and Large Language Models.

\begin{tcolorbox}[
    colback=gray!10,
    colframe=black,
    boxrule=1pt,
    arc=8pt,
    left=12pt, right=12pt, top=10pt, bottom=10pt
]
You are an expert in physics. Solve the following Olympiad-level physics problem step by step. Show the full derivation, including the governing equations, simplifications, and any approximations you invoke. Conclude your response with a line that begins with ``FINAL ANSWER:'' and contains only the final result.
\end{tcolorbox}

Each judge receives the model's full response verbatim and identifies the final result as part of its grading; we do not perform programmatic answer extraction. Multi-line, piecewise, and multi-part answers are handled by the judges directly, with a \emph{partial} verdict available for problems where only some sub-parts are correct.

\subsection{Prompt for Answer Evaluation}

Our evaluation is conducted using three LLM judges: GPT-5.2, Claude-Opus-4.6, and Gemini-3-Pro. The judges use a unified prompt template that takes the reference answer, the reference solution, and the model's full response as input, and returns a structured verdict. For multi-part problems, the judge can return a \emph{partially correct} verdict when some sub-parts are correct and others are not. The full judge prompt is shown in the box below.

\begin{tcolorbox}[
    colback=gray!10,
    colframe=black,
    boxrule=0.5pt,
    arc=8pt,
    left=12pt, right=12pt, top=10pt, bottom=10pt
]
You are an expert physics solution judge. Given a student's response and the reference materials for a physics problem, determine whether the student's final answer is correct.

\medskip
\textbf{Reference Answer}\\
\{reference\}

\medskip
\textbf{Reference Solution}\\
\{solution\}

\medskip
\textbf{Student's Response}\\
\{response\}

\medskip
\textbf{Judging Protocol}

Correctness is decided on the student's final result(s). When the reference answer is empty, extract the correct result from the reference solution.

\begin{enumerate}
    \item Numerical answers are accepted within a 5\% relative tolerance, unless the problem requires an exact integer or rational value.
    \item Symbolic and algebraic answers are accepted in any mathematically equivalent form (e.g., reordered variables, with or without simplification).
    \item For multi-part problems, return ``correct'' only if every sub-part is correct; if some sub-parts are correct and others are not, return ``partially\_correct''.
    \item Disregard differences in units notation, formatting, and intermediate steps. Only the final result is evaluated.
\end{enumerate}

\medskip
Return a single JSON object (no markdown fences) in exactly this schema:

\medskip
\texttt{\{"verdict": "correct" | "partially\_correct" | "incorrect", "reasoning": "<one- to three-sentence explanation>"\}}
\end{tcolorbox}

\section{Scoring Protocol Details}
\label{app:scoring}
\subsection{Answer Score}
The answer score measures whether the model's final answer matches the reference answer. Because \textsc{PhysElite} contains symbolic expressions, numerical values, qualitative conclusions, and equations, scoring allows mathematically equivalent forms when they express the same physical result. Numerical answers are checked with their units and constants when these are part of the required answer.
Answer accuracy is intentionally strict. A response can receive partial credit through the process score even when the final answer is wrong, but it is counted as answer-correct only when the final result is equivalent to the reference.
\subsection{Process Score}
The process score evaluates the derivation. Each model response is first decomposed into key reasoning steps. Each step is then graded as:
\begin{itemize}
    \item \textbf{1.0}: the step is physically and mathematically correct.
    \item \textbf{0.5}: the step uses the right idea but contains a local slip, missing condition, or incomplete execution.
    \item \textbf{0.0}: the step is physically wrong, mathematically invalid, or irrelevant to the problem.
\end{itemize}
We grade the model's own derivation rather than forcing it to align with the reference solution, since physics problems often admit multiple valid solution paths.

\subsection{Judge Ensemble}

Each response is scored independently by three LLM judges: GPT-5.2, Claude-Opus-4.6, and Gemini 3-Pro. The final score is the mean of the three judge scores. The judge prompt instructs each model to identify the major reasoning steps, evaluate each step against the problem statement and reference answer, and avoid penalizing alternative valid derivations.
Human validation of the process scores is reported below.

\subsection{Human Validation of Process Scores}
\label{app:human_validation}

We validated the process-scoring protocol on 800 model responses graded by human experts. A second expert independently graded 400 of these responses. We compared the mean score from the three LLM judges with the human scores on a 0--1 scale.

Pairwise agreement rates were 0.87 between GPT-5.2 and Gemini-3-Pro, 0.81 between GPT-5.2 and Claude-Opus-4.6, and 0.85 between Claude-Opus-4.6 and Gemini-3-Pro. Agreement between the two human experts was 0.95. Table~\ref{tab:human_validation_mae} reports the mean absolute error (MAE) between the averaged LLM-judge scores and the human scores, grouped by subject, answer type, and language.

\begin{table}[htbp]
\centering
\caption{MAE between averaged LLM-judge scores and human scores on the 800-response validation set. Scores are on a 0--1 scale.}
\label{tab:human_validation_mae}
\begin{tabular}{llr}
\toprule
\textbf{Group} & \textbf{Category} & \textbf{MAE} \\
\midrule
Subject & Mechanics & 0.06 \\
Subject & Electromagnetism & 0.08 \\
Subject & Optics & 0.13 \\
Subject & Thermodynamics & 0.09 \\
Subject & Modern Physics & 0.12 \\
\midrule
Answer type & Symbolic expressions & 0.09 \\
Answer type & Numerical values & 0.07 \\
Answer type & Qualitative conclusions & 0.12 \\
Answer type & Equations & 0.08 \\
\midrule
Language & Chinese & 0.09 \\
Language & English & 0.09 \\
\bottomrule
\end{tabular}
\end{table}

\paragraph{Grading disagreement.}
One disagreement occurred on Problem 5801, a Chinese thermodynamics problem answered by Dolphin-Mistral-24B. Two identical balls, initially at temperatures $T_0$ and $2T_0$, are brought into thermal contact in an adiabatic system. The response correctly obtained the final temperature $3T_0/2$ and the total entropy change $mc\ln(9/8)$. Both the LLM judge and the human grader accepted the final answer.

The disagreement concerned the entropy calculation in an intermediate step. The response used $\Delta S=mc\ln(T_f/T_i)$ for each ball. GPT-5.2 assigned 0.5 because the response did not explain the reversible path used to calculate entropy change. The human expert assigned 1.0 because the formula already follows from integrating $dQ/T$ along a reversible path and can be applied using the initial and final states. The graders differed on how much explanation this step required.

\section{Additional Experimental Analysis}
\label{app:results}
\subsection{Model Groups}

Table~\ref{tab:app_model_groups} summarizes the evaluated model groups. The table mirrors the setting in the main experiments and makes explicit which systems are treated as reasoning models and which systems are evaluated as text-only models.

\begin{table}[h]
\centering
\caption{Model grouping used in the evaluation.}
\label{tab:app_model_groups}
\renewcommand{\arraystretch}{1.1}
\begin{tabular}{llll}
\toprule
\textbf{Group} & \textbf{Model} & \textbf{Reasoning Mode} & \textbf{Input Modality} \\
\midrule
\multirow{6}{*}{Closed-source} 
  & Grok-4.2                & System-2 & Text+image \\
  & o3-mini                 & System-2 & Text-only \\
  & GPT-5.2                 & System-2 & Text+image \\
  & Gemini-3-Pro   & System-2 & Text+image \\
  & Kimi-K2-Thinking        & System-2 & Text+image \\
  & Gemini-2.5              & System-2 & Text+image \\
  & Claude-Opus-4.6         & System-1 & Text+image \\
  & Claude-Sonnet-4.5       & System-1 & Text+image \\
  & Qwen-VL-Max             & System-1 & Text+image \\
  & GPT-4o                  & System-1 & Text+image \\
\midrule
\multirow{6}{*}{Open-source} 
  & Qwen3-VL-235B-A22B      & System-1 & Text+image \\
  & Qwen3-VL-32B            & System-1 & Text+image \\
  & Qwen3-VL-8B             & System-1 & Text+image \\
  & Qwen2.5-VL-72B          & System-1 & Text+image \\
  & Dolphin-Mistral-24B     & System-1 & Text+image \\
  & Qwen2.5-VL-7B           & System-1 & Text+image \\
  & DeepSeek-V3             & System-1 & Text-only \\
  & LLaMA-3.1-70B           & System-1 & Text-only \\
\bottomrule
\end{tabular}
\end{table}

\subsection{Language Effects}
Each problem is evaluated in Chinese and English. The main results indicate that most models score higher in English, while Qwen3-VL-235B-A22B is the main exception. This pattern suggests that bilingual evaluation is not merely a translation convenience: it reveals language-specific differences in reasoning, notation handling, and physics expression. The Chinese setting is especially important for \textsc{PhysElite}, because the source problems come from Chinese Olympiad training materials.
\subsection{Difficulty Effects}
The difficulty analysis in the main paper groups problems into Easy (1--2), Medium (3--5), and Hard (6--7). Many models show a U-shaped trend, with Medium problems producing lower scores than the Hard subset. We interpret this as evidence that nominal difficulty alone is insufficient for benchmark interpretation. Hard problems may align more closely with official Olympiad-style templates encountered during post-training, while medium daily-practice problems can be less familiar to current models.
\subsection{Sub-Discipline Effects}
Optics is the most consistent weak point across models. Setup failures are most frequent in optics and least frequent in thermodynamics among the top closed-source systems. This suggests that current MLLMs struggle with the geometric and phase-sensitive aspects of optical reasoning, including ray tracing, interference, and polarization. Mechanics and electromagnetism remain difficult because they often require multi-condition modeling, but their patterns appear more familiar to frontier models.

\subsection{Tool-Assisted Evaluation}
\label{app:tool_assisted}

We conducted a tool-assisted experiment with GPT-4o and Claude-Opus-4.6 on a subset of 400 problems. The subset covered different subjects, difficulty levels, and answer types, with numerical and equation answers sampled more frequently. Both models were evaluated in Chinese and English, and the reported results average the two languages.

Each model had access to a sandboxed Python interpreter with SymPy, NumPy, and SciPy installed. Network access was disabled, and each problem allowed at most three interpreter calls. Prompts and decoding settings matched the main evaluation. We excluded the judge from the same family as the evaluated model and recomputed the no-tool baseline on the same subset with the same grading configuration.

GPT-4o called the interpreter on 44\% of problems. Its answer score increased by 3.5 percentage points and its process score increased by 1.0 point. Claude-Opus-4.6 called the interpreter on 98\% of problems, with increases of 15.0 percentage points in answer score and 9.3 points in process score. Table~\ref{tab:tool_assisted} summarizes these results.

\begin{table}[htbp]
\centering
\caption{Tool-assisted evaluation on 400 problems, averaged across Chinese and English. Gains are relative to the no-tool baseline on the same subset and grading configuration. Answer gains are in percentage points; process gains are in points on the 0--100 scale.}
\label{tab:tool_assisted}
\begin{tabular}{lrrr}
\toprule
\textbf{Model} & \textbf{Tool-call rate} & \textbf{Answer gain} & \textbf{Process gain} \\
\midrule
GPT-4o & 44\% & +3.5 & +1.0 \\
Claude-Opus-4.6 & 98\% & +15.0 & +9.3 \\
\bottomrule
\end{tabular}
\end{table}

\section{Extended Error Analysis}

\label{app:error}

\subsection{Error Taxonomy}
\begin{figure}
    \centering
    \includegraphics[width=1\linewidth]{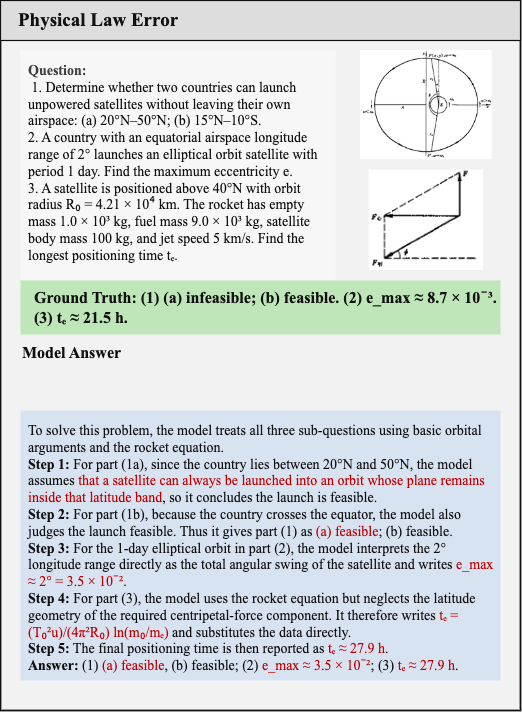}
    \caption{Physcical Law Error}
    \label{fig:ple}
\end{figure}

\begin{figure}
    \centering
    \includegraphics[width=1\linewidth]{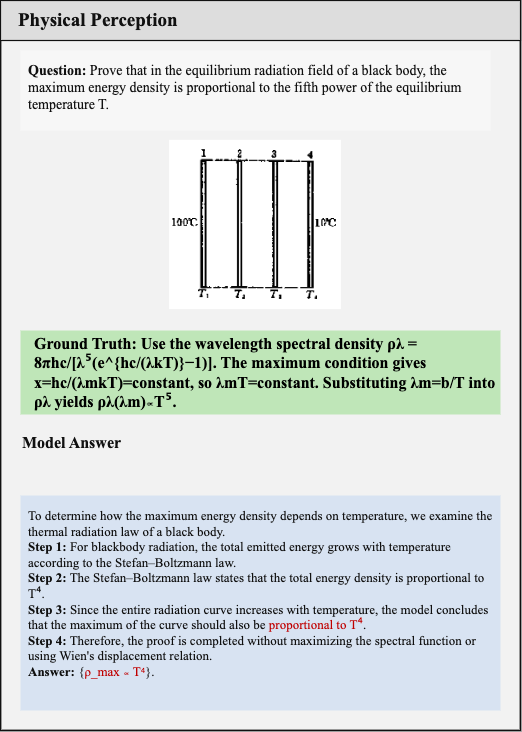}
    \caption{Physical Perception Error}
    \label{fig:ppe}
\end{figure}

\begin{figure}
    \centering
    \includegraphics[width=1\linewidth]{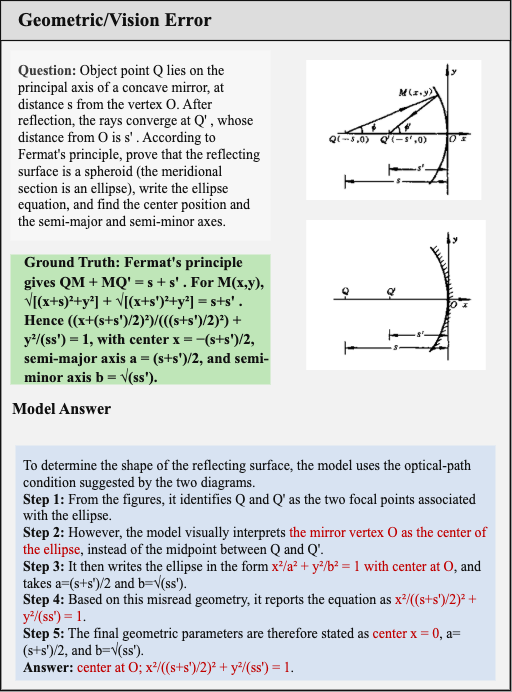}
    \caption{Geometric/Vision Perception}
    \label{fig:gvp}
\end{figure}

\begin{figure}
    \centering
    \includegraphics[width=1\linewidth]{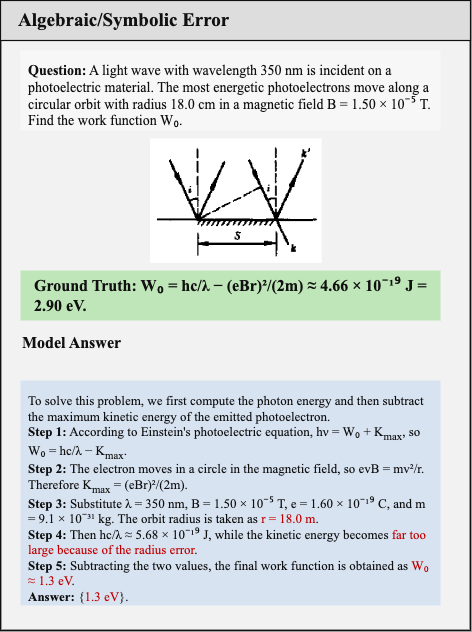}
    \caption{Algebraic/Symbolic Error}
    \label{fig:ase}
\end{figure}

To delve into the failure cases of models, we detailed six typical error types in Table~\ref{tab:app_error_taxonomy} Furthermore, to facilitate a better understanding of each error type, we provide examples of each error made by
Claude-Opus-4.6 from Figure~\ref{fig:ple} to Figure~\ref{fig:ase}.

\begin{table}[h]
\centering
\caption{Detailed Descriptions of Error Types.}
\label{tab:app_error_taxonomy}
\renewcommand{\arraystretch}{1.5}
\begin{tabularx}{\textwidth}{@{}c|X@{}}
\toprule
\textbf{Error Type} & \multicolumn{1}{c}{\textbf{Explanation}} \\
\midrule
Physical-perception error & The model misreads the physical setup, diagram, variables, constraints, or stated conditions. \\
\midrule
Physical-law error        & The model selects an inappropriate physical principle or applies a law outside its valid conditions. \\
\midrule
Algebraic/symbolic error  & The physical setup is mostly correct, but the derivation contains symbolic manipulation, sign, or formula errors. \\
\midrule
Incomplete derivation     & The model starts correctly but stops before all required quantities or sub-cases are solved. \\
\midrule
Geometric/visual error    & The model makes an incorrect geometric inference from a diagram, configuration, angle, path, or spatial relation. \\
\midrule
Numerical mistake         & The model makes an arithmetic, substitution, or unit-conversion error after the correct setup is available. \\
\bottomrule
\end{tabularx}
\end{table}

\subsection{Additional Case Studies}
\label{app:additional_case_studies}

We examine one successful response and two failed responses from the existing evaluation outputs. Chinese problem statements and responses are summarized in English. The step grades below are taken from the GPT-5.2 judging records. Each case reports the answer verdict and the mean of its listed step grades.

\subsubsection{Successful Reasoning: Problem 4270}
\label{app:case_4270}

\textbf{Model and result.} Claude-Opus-4.6; Chinese; difficulty 5; Modern Physics and Electromagnetism. The answer verdict is correct, and the mean step score is 1.00.

\textbf{Problem.} A proton in the Tevatron reaches an energy of $1.00\times10^6\,\mathrm{MeV}$ on a circular orbit with maximum circumference $L_{\max}=6436\,\mathrm{m}$. The problem asks for the minimum magnetic field and the minimum incident kinetic energy for the reaction $p+p\rightarrow p+p+p+\bar p$ on a stationary proton target. It also asks whether the Tevatron beam can produce antiprotons. The proton rest energy is $m_pc^2=938.3\,\mathrm{MeV}$.

\textbf{Model reasoning.} The response relates momentum to the magnetic field through $p=eBr$ and takes the largest orbit radius. It then uses the relativistic energy--momentum relation:
\begin{align*}
r_{\max} &= \frac{L_{\max}}{2\pi}, &
pc &= \sqrt{E^2-(m_pc^2)^2}, \\
B_{\min} &= \frac{2\pi p}{eL_{\max}} \approx 3.26\,\mathrm{T}.
\end{align*}
For antiproton production, the response writes the invariant energy as
\[
s=(E_1+m_pc^2)^2-(p_1c)^2.
\]
At threshold, the final particles are at rest in the center-of-mass frame, giving $s=(4m_pc^2)^2$. Combining this with $E_1^2-(p_1c)^2=(m_pc^2)^2$ gives $E_1=7m_pc^2$ and
\[
T_{\min}=E_1-m_pc^2=6m_pc^2\approx5630\,\mathrm{MeV}.
\]
The response compares this threshold with the Tevatron beam energy and correctly concludes that antiproton production is possible. All six steps receive 1.0, as shown in Table~\ref{tab:case_4270_steps}.

\begin{table}[htbp]
\centering
\caption{Step grades for Problem 4270.}
\label{tab:case_4270_steps}
\begin{tabularx}{\linewidth}{@{}cXr@{}}
\toprule
\textbf{Step} & \textbf{Reasoning} & \textbf{Grade} \\
\midrule
1 & Use $p=eBr$ and obtain $r_{\max}=L_{\max}/(2\pi)$. & 1.0 \\
2 & Obtain the momentum from the relativistic energy--momentum relation. & 1.0 \\
3 & Convert units and calculate $B_{\min}\approx3.26\,\mathrm{T}$. & 1.0 \\
4 & Set up the invariant-energy equation for the production threshold. & 1.0 \\
5 & Derive $E_1=7m_pc^2$ and $T_{\min}=6m_pc^2$. & 1.0 \\
6 & Compare the threshold with the Tevatron beam energy. & 1.0 \\
\bottomrule
\end{tabularx}
\end{table}

\subsubsection{Visual-Perception Error: Problem 4716}
\label{app:case_4716}

\textbf{Model and result.} Claude-Opus-4.6; Chinese; difficulty 4; Optics. The answer verdict is incorrect, and the mean step score is 0.70.

\textbf{Problem.} A thick glass plate is coated with a transparent film of refractive index $n\approx1.3$. The problem provides its transmission spectrum at normal incidence and asks for the film thickness. The reference answer is approximately $50\,\mu\mathrm{m}$.

\textbf{Model reasoning.} The response identifies thin-film interference and correctly derives the transmission maxima and their frequency spacing:
\[
\nu_m=\frac{mc}{2nd},\qquad
\Delta\nu=\frac{c}{2nd},\qquad
d=\frac{c}{2n\Delta\nu}.
\]
The first error occurs when the response reads the spectrum. It estimates six to seven oscillation periods within a $60\,\mathrm{THz}$ interval and uses $\Delta\nu\approx10\,\mathrm{THz}$. The reference solution counts about 34 maxima over $80\,\mathrm{THz}$, giving a spacing of approximately $2.35\,\mathrm{THz}$. Substituting the model's spacing gives
\[
d=\frac{3.00\times10^8}{2\times1.3\times10\times10^{12}}\,\mathrm{m}
\approx11.5\,\mu\mathrm{m}.
\]
Table~\ref{tab:case_4716_steps} locates the error at the spectrum-reading step. The final substitution receives partial credit because the formula and unit conversion are correct, while the frequency spacing leads to the wrong thickness.

\begin{table}[htbp]
\centering
\caption{Step grades for Problem 4716.}
\label{tab:case_4716_steps}
\begin{tabularx}{\linewidth}{@{}cXr@{}}
\toprule
\textbf{Step} & \textbf{Reasoning} & \textbf{Grade} \\
\midrule
1 & Identify thin-film interference and the transmission maxima. & 1.0 \\
2 & Derive $\nu_m=mc/(2nd)$. & 1.0 \\
3 & Derive $\Delta\nu=c/(2nd)$. & 1.0 \\
4 & Read the fringe spacing: the estimate of $10\,\mathrm{THz}$ is too large. & 0.0 \\
5 & Substitute the spacing and calculate $d\approx11.5\,\mu\mathrm{m}$. & 0.5 \\
\bottomrule
\end{tabularx}
\end{table}

\subsubsection{Algebraic Error: Problem 4337}
\label{app:case_4337}

\textbf{Model and result.} Qwen2.5-VL-72B; English; difficulty 4; Thermodynamics. The answer verdict is partially correct, and the mean step score is 0.50.

\textbf{Problem.} An ideal diatomic gas with $n$ moles and molar heat capacity $C_{V,m}=5R/2$ follows a cycle. The path from $A(V_0/2,P_0)$ to $B(V_0,P_0/2)$ is a straight line, and the return path is isothermal. The problem asks for the maximum temperature along $AB$, the process molar heat capacity, the transition between heat absorption and release, the net heat absorbed along $AB$, and the work done over the cycle.

\textbf{Model reasoning.} The response correctly obtains the pressure and temperature along the straight-line path:
\[
P(V)=\frac{3P_0}{2}-\frac{P_0}{V_0}V,\qquad
T(V)=\frac{P_0}{nR}\left(\frac{3V}{2}-\frac{V^2}{V_0}\right).
\]
It correctly locates the temperature maximum at $V=3V_0/4$ and obtains $T_{\max}=9P_0V_0/(16nR)$. The first two steps receive 1.0.

The third step applies $dQ=nC_{V,m}\,dT+P\,dV$. When simplifying the constant terms, the response writes $27P_0/4$ in place of $21P_0/4$. Its expression and the reference expression are
\begin{align*}
\left(\frac{dQ}{dV}\right)_{\mathrm{model}}
&=-\frac{6P_0}{V_0}V+\frac{27P_0}{4}, \\
\left(\frac{dQ}{dV}\right)_{\mathrm{reference}}
&=-\frac{6P_0}{V_0}V+\frac{21P_0}{4}.
\end{align*}
This error gives a transition volume of $9V_0/8$, outside the interval $[V_0/2,V_0]$. The reference transition is at $7V_0/8$, with temperature $35P_0V_0/(64nR)$. The response also leaves the requested expression for the process molar heat capacity unspecified.

Integrating the incorrect heat expression gives $Q_{AB}=9P_0V_0/8$; the reference value is $3P_0V_0/8$. The response correctly computes the straight-line work as $W_{AB}=3P_0V_0/8$. Its final cycle-work expression contains an undefined temperature $T_0$ in the term $-nRT_0\ln2$. The reference cycle work is
\[
W_{\mathrm{cycle}}=\left(\frac{3}{8}-\frac{\ln2}{2}\right)P_0V_0.
\]
The final step receives 0.5. Table~\ref{tab:case_4337_steps} records the correct early steps and the effects of the coefficient error on the later results.

\begin{table}[htbp]
\centering
\caption{Step grades for Problem 4337.}
\label{tab:case_4337_steps}
\begin{tabularx}{\linewidth}{@{}cXr@{}}
\toprule
\textbf{Step} & \textbf{Reasoning} & \textbf{Grade} \\
\midrule
1 & Derive the straight-line equation $P(V)$ from the endpoints. & 1.0 \\
2 & Derive $T(V)$ and calculate its maximum. & 1.0 \\
3 & Determine the molar heat capacity and heat-flow transition; the coefficient error gives the wrong transition. & 0.0 \\
4 & Integrate the heat expression; the result inherits the coefficient error. & 0.0 \\
5 & Calculate cycle work; the straight-line work is correct, but the isothermal term is left in undefined parameters. & 0.5 \\
\bottomrule
\end{tabularx}
\end{table}

\end{document}